\documentclass[sigconf]{acmart}

\AtBeginDocument{%
  }

\setcopyright{acmcopyright}
\copyrightyear{2023}
\acmYear{2023}
\setcopyright{rightsretained}
\acmConference[FAccT '23]{2023 ACM Conference on Fairness, Accountability, and Transparency}{June 12--15, 2023}{Chicago, IL, USA}
\acmBooktitle{2023 ACM Conference on Fairness, Accountability, and Transparency (FAccT '23), June 12--15, 2023, Chicago, IL, USA}
\acmDOI{10.1145/3593013.3594111}
\acmISBN{979-8-4007-0192-4/23/06}

\usepackage{subcaption}
\usepackage{graphicx}
\usepackage{float}

\begin{document}

\title{Capturing Humans’ Mental Models of AI: An Item Response Theory Approach}

\author{Markelle Kelly}
\email{kmarke@uci.edu}
\affiliation{%
  \institution{Department of Computer Science\\University of California, Irvine}
  \city{Irvine}
  \state{California}
  \country{USA}
}

\author{Aakriti Kumar}
\email{aakritik@uci.edu}
\affiliation{%
  \institution{Department of Cognitive Sciences\\ University of California, Irvine}
  \city{Irvine}
  \state{California}
  \country{USA}
}

\author{Padhraic Smyth}
\email{smyth@ics.uci.edu}
\affiliation{%
  \institution{Department of Computer Science\\ University of California, Irvine}
  \city{Irvine}
  \state{California}
  \country{USA}
}

\author{Mark Steyvers}
\email{mark.steyvers@uci.edu}
\affiliation{%
  \institution{Department of Cognitive Sciences\\ University of California, Irvine}
  \city{Irvine}
  \state{California}
  \country{USA}
}

\renewcommand{\shortauthors}{Kelly et al.}

\begin{abstract}
Improving our understanding of how humans perceive AI teammates is an important foundation for our general understanding of human-AI teams.
Extending relevant work from cognitive science, we propose a framework based on item response theory for modeling these perceptions. We apply this framework to real-world experiments, in which each participant works alongside another person or an AI agent in a question-answering setting, repeatedly assessing their teammates' performance. Using this experimental data, we demonstrate the use of our framework for testing research questions about people's perceptions of both AI agents and other people. We contrast mental models of AI teammates with those of human teammates as we characterize the dimensionality of these mental models, their development over time, and the influence of the participants' own self-perception. 
Our results indicate that people expect AI agents' performance to be significantly better on average than the performance of other humans, with less variation across different types of problems. We conclude with a discussion of the implications of these findings for human-AI interaction.
\end{abstract}



\maketitle

\section{Introduction}
With the recent rapid growth in interest in applying AI approaches\footnote{We will use the term ``AI" in this paper to refer to the broad spectrum of techniques that are currently (in 2023) referred to as ``AI,'' including models built using machine learning in particular.} to a wide variety of decision and prediction problems, there is an increasing realization that hybrid human-AI teams will be an important component of how AI will be deployed in practice \cite{kamar2016directions,lai2021science}. 
A decision making process that includes humans and AI can, ideally, benefit from the strengths of each \cite{holstein2021k12,wang2016deep,kamar2012combining}. Humans can act as a safeguard for unpredictable or undesirable behavior in AI algorithms, and can incorporate the type of contextual information and common sense reasoning that AI often lacks \cite{de2020case,
cheng2022welfare,chandler2022applicability}. 
Conversely, the use of AI in prediction and decision making enables the processing of more complex patterns and greater volumes of data than humans alone can accommodate, for example in tedious and time-consuming work such as fact-checking \cite{la2022hybrid} and in high-stakes decision making such as diagnostic medical imaging \cite{Tschandl2020yv}.

One important goal for human-AI systems is \textit{complementarity}---achieving better performance than either the human(s) or the AI agent(s) acting independently \cite{bansal2021does,hemmer2021human,donahue2022unfairness}. A significant body of literature has developed that aims to understand and achieve complementarity across a variety of hybrid human-AI settings \cite{wilder2020complement,pmlr-v151-bordt22a,rastogi2022unifying,steyvers2022bayesian}. One of the lessons learned from this work is that achieving complementarity is complex in practice; for instance, AI agents that exhibit high performance (e.g., in terms of prediction accuracy) on their own can actually harm overall team performance if their behavior is unpredictable for humans \cite{bansal2019updates, bansal2021teamwork}. Thus, understanding humans' expectations of AI is essential for optimizing team performance, and recent work in this area has called for a better understanding of how humans perceive AI \cite{lai2021science,steyvers_kumar_2022}. 

In particular, an important (and somewhat under-studied) topic in this context is humans' mental models of their AI teammates. If a person is deciding whether or not to take the advice of an AI agent, they are likely to make better decisions if they can accurately perceive its strengths and weaknesses, that is, if they have developed an accurate mental model of the agent \citep{bansal2019beyond}. For example, an AI agent could have particular ``blindspots'' in terms of its expertise, even when the agent's overall performance on a particular task is comparable to or exceeds that of a human \cite{attenberg2015beat,d2022spotlight}.
In general, a better understanding of human mental models of AI agents can help predict phenomena such as humans' development of appropriate trust in AI agents, human behavior in deferring to an AI agent, and how a human-AI team will function overall. 

In this paper, we aim to improve our understanding of human mental models of AI agents by building on prior theoretical and empirical work in cognitive science that has analyzed how people form mental models of other people.
To this end, we present a general framework for modeling human mental models of AI agents based on item response theory (IRT). In the traditional IRT approach, given an agent's performance on problem sets involving a particular task, the IRT framework is used to estimate both problem set difficulties and agent abilities.  
We build on this traditional IRT methodology to propose a new framework that models human mental models both of themselves and of others, in terms of perceived abilities and perceived problem difficulties.
This enables us to make comparative predictions about humans' perceptions of AI agents, in the context of their perceptions of themselves and of other people.

To highlight the use of this framework, we conduct experiments in the context of question-answering where participants work alongside either another person or an AI agent (in the form of a large language model). Participants estimate the performance of their counterpart throughout the experiment, allowing us to make inferences about participants' perceptions (mental models) of the abilities of their counterparts. In our analysis, we focus on two sets of research questions that have not yet been explored in the literature:
\begin{enumerate}
\item {\bf Multidimensionality of Mental Models:} What is the dimensionality of humans' mental models in the context of assessing task performance of other agents? In particular, do humans' mental models of others capture multiple different abilities or areas of expertise, or do they estimate a single notion of ability, a ``general intelligence''? We investigate the specific structure of these perceptions of ability, in particular, the correlations between different abilities, and how our findings differ between mental models of other humans and those of AI agents.
\item {\bf Role of Self-Perception:} What role does a person's self-perception play as they develop a mental model of another agent? For example, do people estimate how another agent's abilties differ from their own? We consider multiple potential relationships between (a) humans' perceptions of their own abilities and experienced problem difficulties and (b) their perceptions of other agents' abilities and problem difficulties. Again, we explore how these relationships differ between mental models of other humans and AI agents.
\end{enumerate}
To address these questions we make two main contributions in this paper. First, we present an extensive experimental dataset involving humans and AI in a question-answering context. This data directly captures human perceptions of self- and other-agent performance, providing insight into questions about humans' mental models of other agents. Second, we
introduce a theoretical model-based framework for directly modeling and analyzing humans' mental models of AI agents, and we use this modeling framework to gain insight into our experimental data.\footnote{All our code and data, including the original trivia questions, are available at \href{https://github.com/markellekelly/AI_mental_models}{https://github.com/markellekelly/AI\_mental\_models}.}

In Section \ref{sec:relatedwork}, we review existing work on mental models and their role in human-AI teams. In Section \ref{sec:experiments}, we describe our experimental setup, and we include empirical findings and data analysis in Section \ref{sec:analysis}. Section \ref{sec:framework} introduces our IRT-based framework, and in Sections \ref{sec:rf1} and \ref{sec:rf2} we present the methodology and results for our two Research Foci---Section \ref{sec:rf1} on the dimensionality of mental models and Section \ref{sec:rf2} on the influence of self-perception. Finally, we discuss key takeaways in Section \ref{sec:discussion} and conclude in Section \ref{sec:conclusion}.

\section{Related Work and Background}
\label{sec:relatedwork}

\subsection{Mental Models and Collaboration} 
In general, \textit{mental models} are simplified representations of the world that people use to process new information and make predictions \cite{Barnes1944-cg,Smyth1994-qo}. 
Our work focuses in particular on mental models of one's self, of other people, and of AI agents, and is informed by foundational work in cognitive science in the areas of \textit{metacognition} \cite{livingston2003metacognition,dunlosky2008metacognition}, \textit{theory of mind} \cite{astington1995theory,frith2005theory}, and \textit{theory of machine} \cite{logg2022psychology}, respectively.

The importance of mental models of other agents has received considerable emphasis in prior work on collaboration, particularly for collaboration among teams of humans. Specifically, the goal of shared mental models (SMMs) \cite{scheutz2017framework,merry2021mental,schelble2022shared} necessitates that team members are aligned in terms of their perceptions of their team, strategy, and the task at hand. Information about the skills and knowledge of a teammate, which we refer to as an ``other mental model'' or OMM, is an important component of these SMMs, promoting effective collaboration \cite{matheieu2000shared,sidera2018theory,miller2019explanation,buehler2020theory,druce2021brittle}. In the context of the focus of this paper, namely hybrid human-AI teams, prior work has found that more accurate perceptions of AI agents tend to result in better team performance \cite{gero2020mental,bansal2021teamwork} and more satisfying interactions for humans \cite{kulesza2012tell}. 

\subsection{Understanding Mental Models of AI Agents}
Given the importance of human perceptions of AI agents in human-AI collaboration, an emerging body of work aims to understand these perceptions \cite{westermanit2020, wangtowards2021,chen2023understanding}. Based on experiments in cooperative game settings, \cite{gero2020mental} delineated three different components of mental models of AI: the agent's \textit{knowledge distribution}, \textit{local behavior}, and \textit{global behavior}. There is evidence that people develop a mental model of an AI agent's \textit{knowledge distribution} based on their own knowledge \cite{lee2005robots}, and that this perceived intelligence can be affected by provided explanations \cite{paleja2021utility}. It has also been shown that humans develop perceptions of AI agents' \textit{global behavior}, but that these perceptions weaken as error boundaries become more complex and stochastic \cite{bansal2019beyond} and can be biased by first impressions of the agent \cite{nourani2021anchoring}. Finally, prior research has shown that people can predict an AI agent's \textit{local behavior}, basing initial estimates on their own abilities \cite{bos2019mental}, and that counterfactual examples can improve these predictions \cite{alipour2021improving}. 

In this paper, we characterize mental models of AI agents in terms of the perceived ability of the agent and the perceived problem difficulty for the agent. While we use perceptions of behavior to estimate these quantities, all three of \cite{gero2020mental}'s components of mental models are relevant. A human's perception of the abilities and problem difficulties for an AI agent could be influenced by how that human perceives the agent's knowledge distribution. Perceived ability and problem difficulty could also be used to predict both local and global behavior.

\subsection{Contributions in the Context of Related Work}
The focus of this paper differs from prior work on mental models of AI agents in two important ways. First, we introduce a framework that directly models OMMs, and thus can describe them in terms of relevant latent variables and test hypotheses about their structure. In contrast, prior work generally 
has not investigated OMMs directly, instead analyzing a proxy such as team performance or humans' predictions of AI behavior (e.g., \cite{bansal2019updates, alipour2021improving}). Second, in our experiments, we collect data on participants' mental models of other people. Directly comparing mental models of AI agents to those of other people provides important context and helps determine how human-AI collaboration relates to general cognitive science research on teaming.

More specifically, our work differs from that of \cite{bansal2019beyond}, who also experimentally investigated the capacity of humans to understand AI agents. 
Their analysis focused on the relationship between the complexity of the AI agent's classification error boundary and the participant's performance on the task. This earlier work differs from the work in this paper in that it did not directly capture or model participants' OMMs, nor did the approach relate OMMs to self-perceptions or mental models of other people. Our work also differs from \cite{nourani2021anchoring}, who investigated experimentally how participants' mental models of an AI agent were affected by their first impressions of the agent. 
Their results demonstrated that people developed more accurate mental models (i.e., had lower error in predicting model performance) when they had a positive first impression. However, the mental models themselves were not analyzed in terms of their structure, and they were not compared to self-perceptions or OMMs of people.

\begin{figure*}[!ht]
    \centering
    \includegraphics[trim=1cm 5cm 1cm 3cm,clip,width=\linewidth]{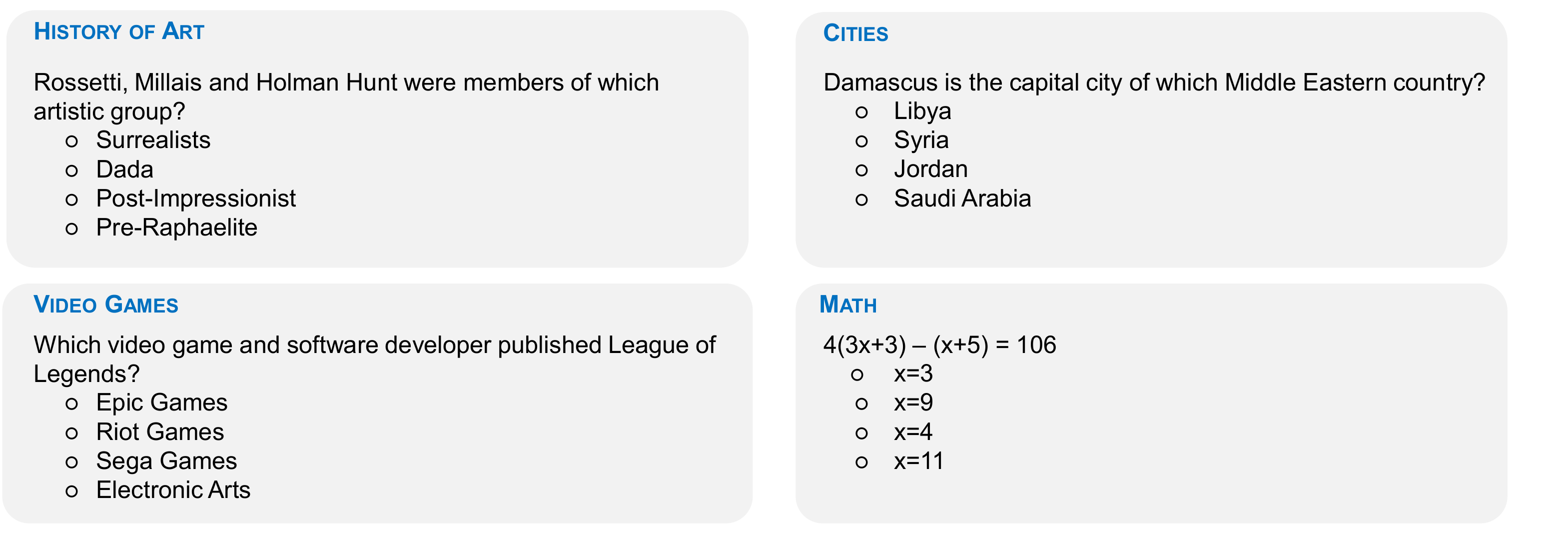}
    \caption{Sample questions from each trivia category.} 
    \label{fig:sampleqs}
\end{figure*}


Finally, there has been recent prior work that has proposed analytic frameworks for human mental models of other humans. In particular, \cite{kumar2023differentiating} developed an IRT framework for understanding these OMMs, which we build upon in this paper. Further, \cite{Westby2022CollectiveII} introduced an AI agent that learns human OMMs of their human teammates for the purpose of improving collective intelligence in a human-human teaming context. This is relevant to the approach we propose in this paper in that it demonstrates how a framework such as ours could be deployed in a team scenario. However, these prior frameworks on modeling OMMs differ from our approach in that they do not investigate mental models of AI agents. 

\section{Experiments}
\label{sec:experiments}
In our experiments, participants complete a multi-category trivia question-answering task, estimating their own performance and the performance of either another human or an AI agent. Experiments were conducted using Amazon Mechanical Turk.

\subsection{Task}
\label{sec:task}
Each participant answered 16 sets of 12 trivia questions. We used a trivia setting because it does not require specialized knowledge (and thus is doable by Mechanical Turk workers), is discriminative (it is very unlikely humans or AI will answer all trivia questions correctly) \cite{graber2019trivia}, and can be broken up into distinct categories, allowing us to directly investigate multiple dimensions of ability.

Using a dataset of trivia questions from \href{https://www.thequestionco.com/}{The Question Company}, we selected four question topics: \textit{History of Art}, \textit{Video Games}, \textit{Cities}, and \textit{Math}. Sample questions for each topic are shown in Figure \ref{fig:sampleqs}. Using preliminary data, we selected these topics to achieve  (1) variation between participants for each topic, (2) variation among topics for each participant, and (3) varying correlations between pairs of topics across participants. 
In addition, we selected topics that we expected people would \textit{perceive} as being different, e.g., requiring different types of knowledge.

\begin{figure*}
    \centering
    \includegraphics[width=0.8\linewidth]{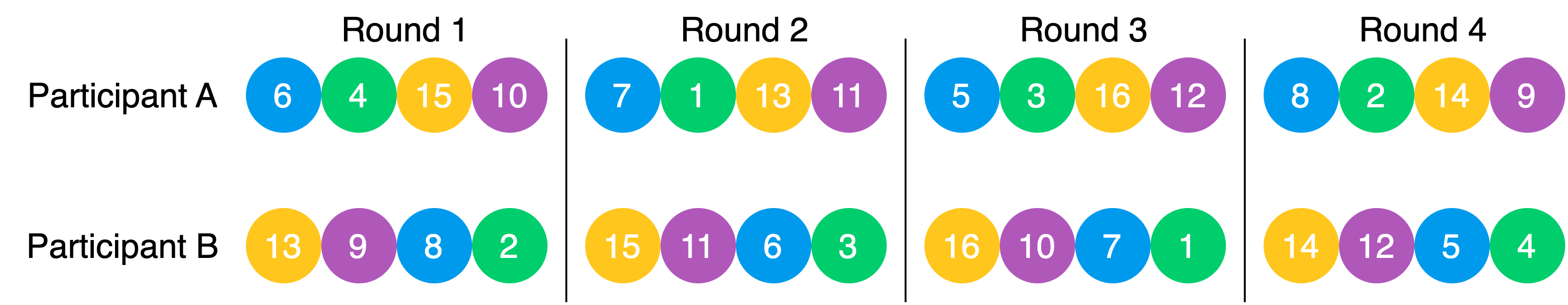} 
    \caption{Example experiment configurations. Each circle represents a problem set, identified by its ID (number) and topic category (color). Each round has one problem set from each category and the order of these categories is consistent across rounds. Both this order and the order of individual problem sets is randomized for each participant.}
    \label{fig:randomization}
\end{figure*}

For each category, we created four problem sets of 12 questions each, for a total of 16 problem sets. Participants were presented with questions in four rounds. Each round included one problem set from each trivia category. For each participant, we randomized the order of questions within each problem set, the order of categories within rounds, and the order of problem sets across rounds. 
Examples of possible problem set orders are shown in Figure \ref{fig:randomization}.

\subsection{Performance Assessment}
After each 12-question problem set, participants were asked to estimate their own performance, to capture self-assessment, and the performance of another agent, to capture other-assessment. Participants were randomly assigned to assess either an ``AI system'' or another person (see Appendix \ref{sec:experimentsetup} for details).

Two-thirds of the participants were randomly selected to receive feedback regarding their performance estimation: after they provided their estimates for a given problem set, they were shown their own actual performance, and the actual performance of the other agent, on that problem set. The remaining one-third of the participants did not receive this feedback. This no feedback group was included to capture prior expectations and to understand the strategies people use to estimate the performance of another agent in the absence of feedback.

\subsection{``Other'' Selection}
To obtain the performance data for the other humans, we first performed a pilot study ($n=34$) using the same 16 12-question problem sets. Based on overall accuracy, we selected the top five (``high accuracy'') and bottom five (``low accuracy'') participants in the pilot study, where each of the top five and bottom five were then used as ``other humans" in the main experiment. Specifically, participants in the main experiment that were assigned to the other human condition were shown performance data from one of these 10 participants. The name of this other human was randomly chosen from a set of ten names drawn from a random name generator (e.g., ``Anna'' or ``Felix'').

To reduce the possibility of performance as a confounding variable, we then matched AI performance with the other human performance on a topic-wise basis. This was done by running several variants of UnifiedQA \cite{khashabi2020unifiedqa} and Zero-shot-CoT \cite{kojima2022large}, which are large language models designed to generalize to a range of tasks. We then chose two models for each topic, one with similar performance to that of the high accuracy humans, and another to match the performance of the low accuracy humans. (Details on the exact model settings used, and the final topic-wise accuracies, can be found in Appendix \ref{sec:experimentsetup}.)

We include agents with both high and low accuracy to improve the generalization of results; mental models could differ depending on whether the other agent has higher or lower performance than the participant.

\subsection{Setup}
203 Amazon Mechanical Turk workers, all located in the U.S., participated in the study, which was conducted in January 2023. Participants could only complete the experiment once (and were disqualified if they had participated in the earlier pilot study). To ensure high-quality participation, workers were required to be AMT Masters and have a 95\% approval rating; they were paid \$7 plus a bonus of up to \$2. These incentive bonuses were based on the participants' other-assessment performance. The experimental protocol was approved by the University of California, Irvine Institutional Review Board.

Participants were assigned to an agent type (other human or AI agent) and agent category (high accuracy, low accuracy, or no feedback). Participants were evenly divided between these six experimental condition combinations. 

\section{Overview and Empirical Analysis of Experimental Data}
\label{sec:analysis}

We begin our analysis with an investigation of initial findings from our experimental data, exploring the participants' perceptions of their own (self) performance, compared to their perceptions of the performance of others (both human and AI).

\begin{figure*}
    \centering  \includegraphics[width=0.5\linewidth]
    {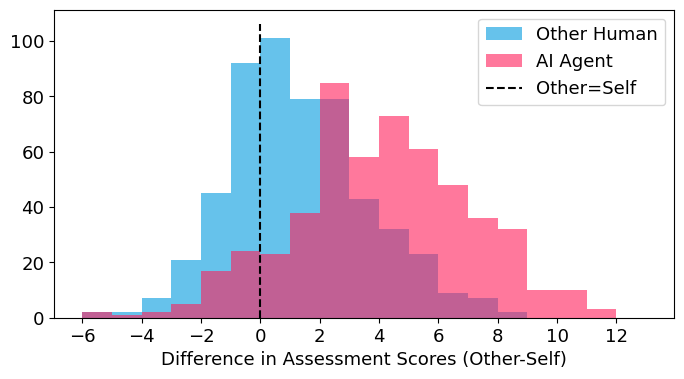}
    \caption{Histograms of differences between other agent (AI or Human) assessments and self assessments in the no feedback condition.}
    \label{fig:nofeedback}
\end{figure*}

\subsection{No Feedback Condition}
We consider first the no feedback experimental condition, examining participants' OMMs when no performance information is available about the other agent. We analyze aggregate results for all 1072 problem set assessments in the no feedback condition. 
Each participant provided a score between 0 and 12 estimating their own performance (self-assessment) and a score estimating the performance of another (human or AI) agent (other-assessment), in terms of how many questions were answered correctly, for each of 16 problem sets. Figure \ref{fig:nofeedback} shows a histogram plot of the differences between self-assessment and other-assessment for both AI agents and other humans. Positive differences indicate that a participant provided a higher score for the other agent than for themselves.

The results in Figure \ref{fig:nofeedback} illustrate a striking difference between participants' assessments of AI agents and their assessments of other humans. The mean difference of the assessed performances of AI agents (relative to self) was $+3.0$ points, compared to $+0.8$ points for other humans (the two means are significantly different, and both are greater than 0, at a p-value threshold of $\alpha=0.01$ under two-sample and one-sample one-tailed t-tests, respectively). 
These results indicate that on average, people believe other agents will perform better than themselves in trivia question answering, and that AI agents will have much higher performance than other humans. Our findings are consistent with previous research indicating that people expect AI to be better at objective tasks, e.g., tasks that involve retrieving factual information, when compared to humans \cite{castelo2019task, logg2017theory}.

\subsection{Feedback Condition}
In order to investigate how participants' OMMs adapt over time, given feedback about the performance of the other agent, we analyze the experimental data for all 2176 problem set assessments under the feedback condition. 
Of interest in this context is how feedback about the other agent (provided after each self-assessment, for 16 problem sets per participant) affects people's perceptions of the performance of the other human or of the other AI agent.


\begin{figure*}
  \begin{subfigure}[b]{0.35\textwidth}
    \includegraphics[width=\textwidth]{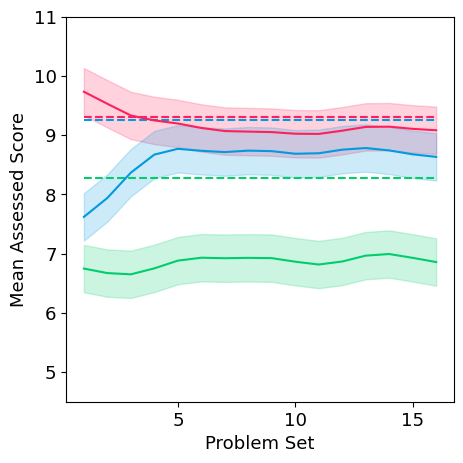}
    \caption{High Accuracy Agents}
  \end{subfigure}\qquad\qquad
  \begin{subfigure}[b]{0.35\textwidth}
    \includegraphics[width=\textwidth]{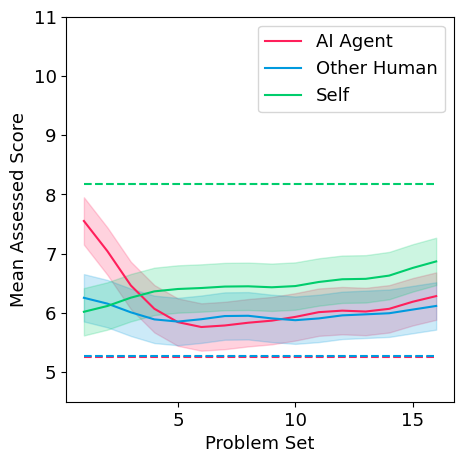}
    \caption{Low Accuracy Agents}
  \end{subfigure}
    \caption{Mean perceived performance of other agents and self at each problem set. The results are separated by other agents with (a) high accuracy and (b) low accuracy. Dashed lines show corresponding values of actual performance for reference (for self, AI agents, and other humans, averaged across all participants and all problem sets in the given experimental condition). Results are smoothed across problem-sets to facilitate visual comparison. }
    \label{fig:withfeedback}
\end{figure*}

Figure \ref{fig:withfeedback} illustrates how participants update their assessments over rounds of feedback, starting from problem set 1 (when no feedback has been provided yet) up to problem set 16 (when feedback has been provided about all previous 15 problem sets). Participants adapt their assessments of other agents (both AI and human, both high and low accuracy) quickly within the first 4 to 6 problem sets and then change relatively slowly after that. 
Even after feedback from 16 problem sets, for the high accuracy agents, participants still systematically assess AI agents as being roughly 0.6 points more accurate than the human agents, even though the two agents were selected to have approximately the same accuracy (dotted lines). For the low accuracy agents there is also a consistent bias in favor of the AI agent, of roughly 0.2 points. 

We also note that feedback appears to have relatively little influence in correcting self-assessed performance (in green).
For example, for the low accuracy agents, after feedback on 16 problem sets, even though the human and AI agents are much worse (-3 points) in terms of actual performance than the average participant (dotted lines), participants predict similar performance for themselves and the low accuracy agents.

In summary, the experimental results show that OMMs are quite different for AI agents and other humans for this question-answering task. People generally expect, a priori, that AI agents outperform humans, and it takes considerable evidence to adjust these expectations.

\section{IRT Framework}
\label{sec:framework}
In this section we briefly outline our theoretical modeling framework for mental models, which we then use in Sections \ref{sec:rf1} and \ref{sec:rf2} to investigate Research Foci 1 and 2 posed in the Introduction. Our framework is based on Item Response Theory (IRT) \cite{fox2010bayesian, van2013handbook} which is widely used in education \cite{bichi2018item} and psychology \cite{thomas2019advances} for modeling observed performance on a task in terms of latent psychological factors. In particular, we extend the hierarchical model of knowledge assessment proposed in \cite{kumar2023differentiating}, which focused on people's mental models of their own knowledge and the knowledge of other humans, but did not investigate mental models of AI agents. 

In a standard IRT setup, we have $V$ problems per problem set $j$, where each problem set $j$ has a latent (unobserved) difficulty $d_j$. For each individual
  $i$ and each problem set $j$, we model the number of items answered correctly $x_{i,j}$, ranging between 0 and $V$, as:
\begin{equation*}
    x_{i,j} = f(\theta_{i,j}) = f(a_i - d_j) 
\end{equation*}
where $a_i$ is the latent ability of individual $i$ and $f$ is a function that noisily converts the latent $\theta_{i,j}$ for individual $i$ and task $j$ into an integer-valued $x_{i,j}$. The equation above represents how the IRT model can simulate or generate data in a forward manner. Given observed data $x_{i,j}$, for multiple participants $i$ and problem sets $j$, we can then make inferences in the reverse direction about the latent abilities $a_i$ and problem difficulties $d_j$ (e.g., using standard Bayesian sampling techniques such as Markov Chain Monte Carlo sampling).

As in \cite{kumar2023differentiating}, our IRT framework analyzes \textit{perceived} performance in addition to actual performance. In our experiments, we ask participants to estimate their own performance on each problem set. We denote this self-perceived data $x_{i,j}^s$, using an $s$ superscript to signify self-assessment, with $x_{i,j}^s = f(\theta_{i,j}^s) = f(a_i^s - d_j^s)$. Participants are also asked to assess the performance of another human or an AI agent, capturing their OMM. We refer to this data with an $o$ superscript, i.e., $x_{i,j}^o = f(a_i^o - d_j^o)$. Thus, using the other-assessment data $x_{i,j}^o$, we can then estimate $a_i^o$, the perceived ability of the other agent, as well as $d_j^o$, the perceived difficulty of problem $j$ for the other agent, all from the perspective of participant $i$. Further details of our modeling framework are provided in Appendix \ref{sec:irtdetails}.

\section{Research Focus 1: Multidimensionality of Mental Models}
\label{sec:rf1}
We next investigate whether people develop multidimensional OMMs, and more specifically, if people develop estimates of multiple different abilities of another agent. In the context of our experiments, we explore whether participants assess the strengths and weaknesses of the other agents across different topics. We are also interested in the correlational structure of these perceived abilities: do people expect other agents to have expertise in specific topics, or do they expect other agents to have more generalized intelligence? 
Finally, we investigate how these mental models develop over time and how these mental models differ between other humans and AI agents.

\subsection{Methods}
We address the question of multidimensional models of ability by comparing
 one-dimensional and multidimensional IRT models for other-assessment on our experimental data. Multidimensional item response theory (MIRT) \cite{reckase1997future,Ackerman1994-lk} models ability as a vector with multiple dimensions. For example, a student's performance on a standardized test might be related to their reading, writing, and mathematical abilities. In our framework, we use a between-items MIRT setup, which assumes that the probability of success for a specific item is affected by only one of the ability dimensions \cite{Sheng2007-mi,HARTIG200957}. 

In particular, we associate each of the four trivia topics (history of art, video games, cities, and math) (see Section \ref{sec:task}) with an ability dimension. Thus, each component of $\mathbf{a}_i^o$ corresponds to the perceived ability of the other agent in a specific trivia category. 
To understand the dimensionality of perceived abilities, we can then compare the one-dimensional and multidimensional models: 

\begin{equation*}
\begin{aligned}[c]
&\text{One-dimensional}\\
&\theta_{i,j}^o = a_i^o - d_j^o\\
\end{aligned}
\qquad\qquad\qquad\qquad
\begin{aligned}[c]
&\text{Multidimensional}\\
&\theta_{i,j}^o =   {a}_{ij}^o - d_j^o  
\end{aligned}
\end{equation*}
where, in the multidimensional model, ${a}_{ij}^o $ is the component of the $k$-dimensional ability vector $\mathbf{a}_{i}^o $ that corresponds to problem $j$.

Further, the MIRT model estimates a $4 \times 4$ matrix $\Sigma^o$ of latent linear correlations between ability dimensions (see details in Appendix \ref{sec:irtdetails}). 
These estimated correlations capture the structure of perceived abilities, allowing us to quantify how participants expect these topic-wise abilities to be related.

\subsection{Results}

\begin{figure*}[!ht]
    \centering
    \begin{subfigure}{0.3\textwidth}  
    \centering
        \includegraphics[width=\linewidth]{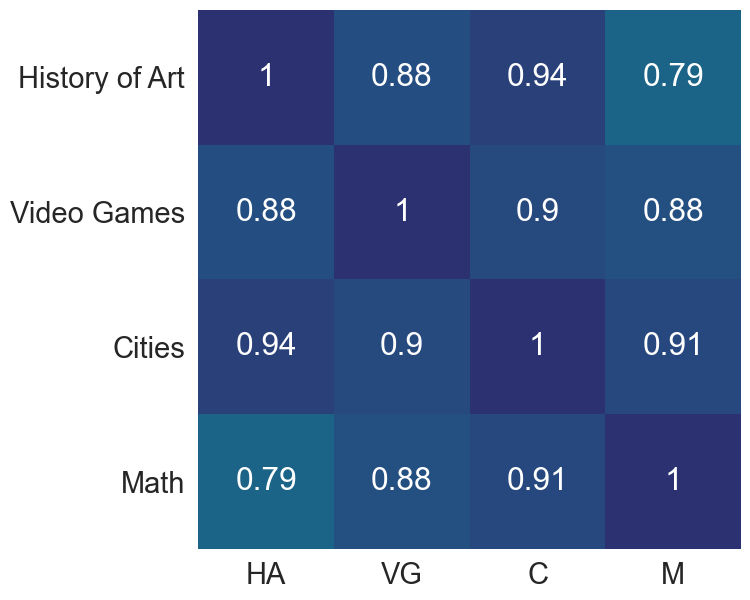}    \caption{AI agents}
        \label{fig:aicorrs}
    \end{subfigure}\qquad%
        \begin{subfigure}{0.3\textwidth}    
    \centering
        \includegraphics[width=\linewidth]{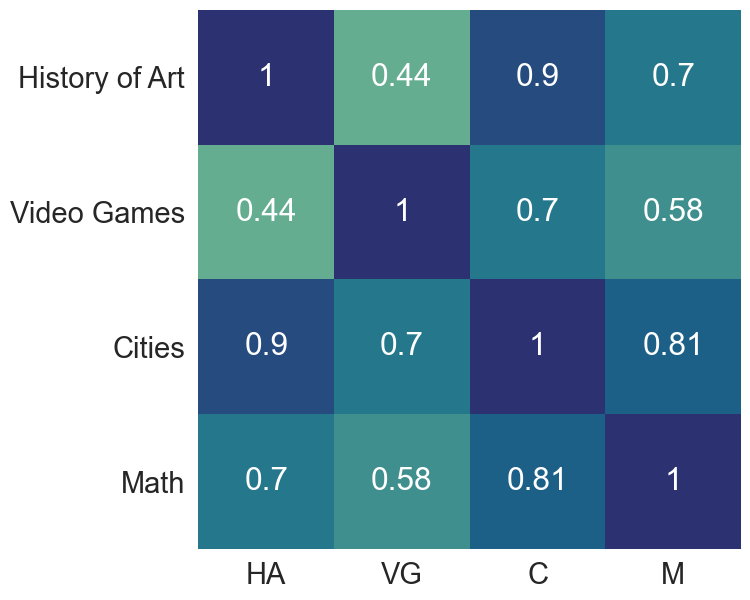}
        \caption{Other humans}
        \label{fig:humancorrs}
    \end{subfigure}\qquad%
    \begin{subfigure}{0.3\textwidth}  
    \centering
        \includegraphics[width=\linewidth]{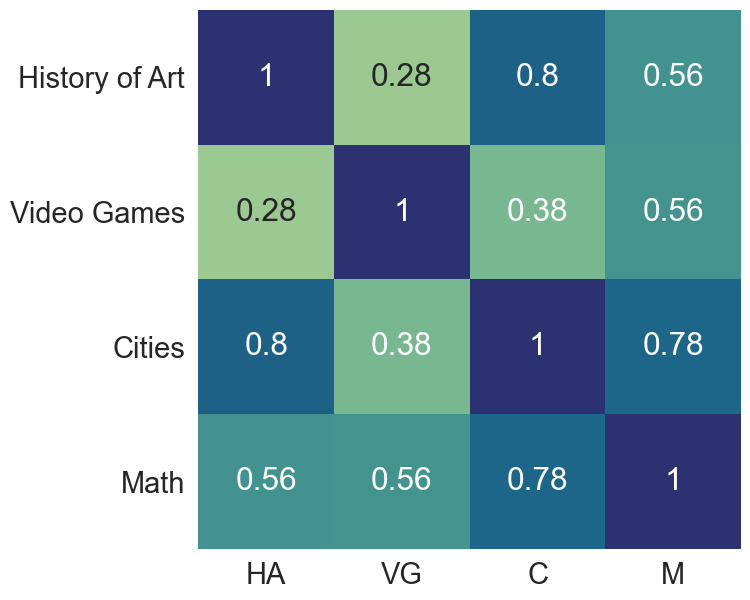}
        \caption{True correlations}
        \label{fig:truecorrs}
    \end{subfigure}%
    \caption{Latent ability correlations for the four trivia categories in the assessment of AI agents (a) and other humans (b), and for the true performance of both humans and AI agents (c). The results are based on the feedback condition.}
    \label{fig:corrs}
\end{figure*}

\subsubsection{Dimensionality of mental models}
\label{sec:dimensionality}
To determine the dimensionality of participants' OMMs, we compare the simpler one-dimensional model with the multidimensional model using three widely-used statistical model selection techniques: (i) held-out log-likelihood, (ii) WAIC score, and (iii) LOO score \cite{vehtari2017practical, luo2017performances}. \footnote{In the main paper, for all model comparisons, we present only the held-out log-likelihood scores for brevity. See Appendix \ref{sec:additionalresults} for additional details; all three scores agree in terms of which models were selected in all model comparisons.} For each participant, we hold out the final four problem sets they completed, and compute the log-likelihood over those four sets. Average held-out log-likelihoods across problem sets and participants are shown in Table \ref{table:multidim}. For reference, throughout the paper we contrast the performance of the IRT models with a discrete uniform baseline on $[0,12]$.

\begin{table}[!ht]
\centering
\caption{Held-out log-likelihood (higher is better) of the baseline, one-dimensional, and multidimensional models for other humans and AI agents, in the feedback condition.}
\begin{tabular}{lll}
           & Humans & AI \\
 \hline\hline
Baseline & -2.56 & -2.56 \\
One-dimensional & -1.81    &  -1.82 \\
Multidimensional & \textbf{-1.74}  & \textbf{-1.72} \\
\end{tabular}
\label{table:multidim}
\end{table}

The held-out log-likelihoods (and WAIC and LOO scores) indicate that the multidimensional model is a better model for both other-human and AI agent assessment. 
These results suggest that people can, and do, develop ideas of another agent's strengths and weaknesses; mental models are not limited to a single ability.

\subsubsection{Latent correlations of ability dimensions}
\label{sec:corrs}
Given that there is evidence that perceptions of others' abilities are multidimensional, we are interested in their specific correlational structure. For instance, people might expect agents to have specific pockets of expertise or to exhibit a more general intelligence \cite{spearman1904general,HUMPHREYS1979105}. 

To this end, we compare the latent correlations between trivia topics for other assessment, shown in Figure \ref{fig:corrs}. Here we include only participants in the feedback condition, capturing differences in structure even when participants have observed the agent's actual performance across these topics. We also include the true correlations across both humans and AI agents. 
 (Note that these true correlations are similar between other humans and AI agents; the correlations split by agent type can be found in Appendix \ref{sec:additionalresults}.)

In summary, we find that the perceived correlations between abilities are significantly higher when assessing AI agents than when assessing other people. This means that participants expect the abilities of AI agents across different trivia categories to be highly correlated with each other, much more so than the correlations across abilities of another person. 
Note that this does not contradict the earlier findings of multidimensionality in OMMs; the perceived abilities per topic are correlated, but they are still distinct from each other.

\subsubsection{Mental models over time}
\label{sec:mdtime}
\begin{figure*}[!ht]
    \centering
    \begin{subfigure}{0.5\textwidth}
    \centering
        \includegraphics[width=0.5\linewidth]{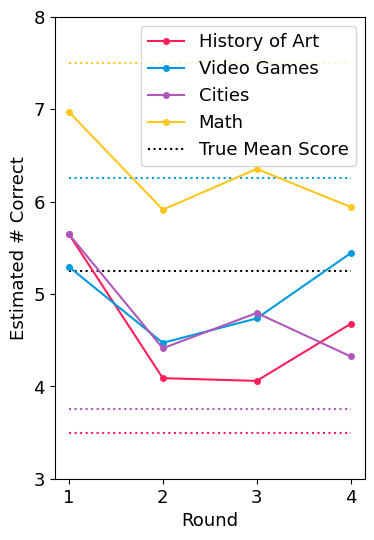}
        \caption{AI agents}
        \label{fig:timeai}
    \end{subfigure}%
    \begin{subfigure}{0.5\textwidth}
    \centering
        \includegraphics[width=0.5\linewidth]{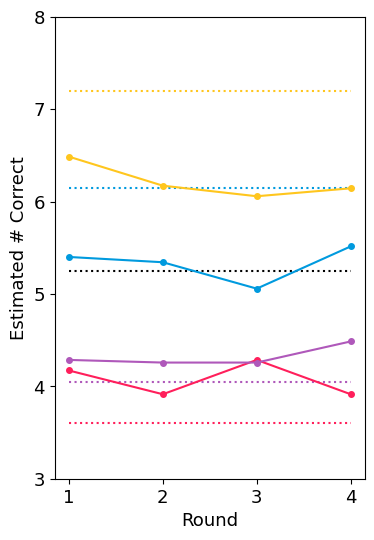}
        \caption{Other humans}
        \label{fig:timehuman}
    \end{subfigure}
    \caption{Average other-assessment for low accuracy agents in the feedback condition. In both figures, the solid lines plot the average other-assessed performance across the four rounds. The average true performances of the other agent are shown as dotted horizontal lines, with the overall average in black and per-topic averages in their respective colors.}
    \label{fig:time}
\end{figure*}

In Section \ref{sec:analysis} we saw that other-assessment (on average) tends to stabilize, without converging to the other agent's true performance, over the first 4 to 6  problem sets, with some continued drift for low accuracy agents. In this section we explore how other-assessment develops at the per-topic level, that is, how multidimensionality develops over time. In Figure \ref{fig:time}, other-assessment data for low accuracy agents is plotted across rounds (where a round consists of 4 problem sets for a participant, one problem set from each topic). (Similar plots for high accuracy agents can be found in Appendix \ref{sec:additionalresults}.)

Figure \ref{fig:time} reflects that participants' estimates of agent performance do not converge to the true agent performance at the per-topic level. Although in Round 4 there are differences, on average, between performance estimates for different topics (i.e. perceptions of ability are multidimensional), the estimates do not appear to have stabilized, and are not spread out enough to match true per-topic performance, particularly for AI agents. Instead, they appear to be anchored by the overall average performance (shown in black).

In line with the findings of Section \ref{sec:corrs}, Figure \ref{fig:time} also reflects that people expect the abilities of AI agents to be highly correlated. Initially, participants expect similar performance across history of art, video games, and cities questions for AI agents (see the cluster of points in Round 1 in Figure \ref{fig:timeai}). For other humans, the estimates are more spread out across topics. After this first round, participants' estimates of AI agent performance exhibit a similar decrease across all four categories---even though initial estimates of math and video games performance were, on average, underestimates. This reflects high perceived correlations between abilities: people expect AI agent scores to be closely related across topics, more so than for other humans (Figure \ref{fig:timehuman}). 

\section{Research Focus 2: Influence of Self-Perception}
\label{sec:rf2}
In this section, we explore the role a person's self-perception plays when forming a mental model of another agent. In particular, we investigate whether the role of self-perception in developing an OMM differs between AI agents and other people, as well as how this changes as more information about the other agent becomes available. 
Developing an understanding of the differences between one’s own capabilities and those of an AI agent is essential for improving cooperation \cite{steyvers_kumar_2022}, and explicitly comparing one's own performance with the performance of an AI agent can promote appropriate selective reliance on the algorithm \cite{liang2022adapting}. 

\subsection{Methods}
We test the fit of three different hierarchical IRT structures connecting the true (``underlying'') performance, self-assessment, and other-assessment data. We assume that self-assessment is (noisily) related to true performance. More specifically, we assume that self-assessed ability is a function of a person's underlying ability, and self-perceived difficulty is a function of the underlying problem difficulty. We then test the relationship between self-assessment and other-assessment latent parameters, following a three-tier hierarchical structure.

\begin{figure*}[!ht]
\centering
    \begin{subfigure}{0.2\textwidth}
    \centering
        \includegraphics[width=0.8\linewidth]{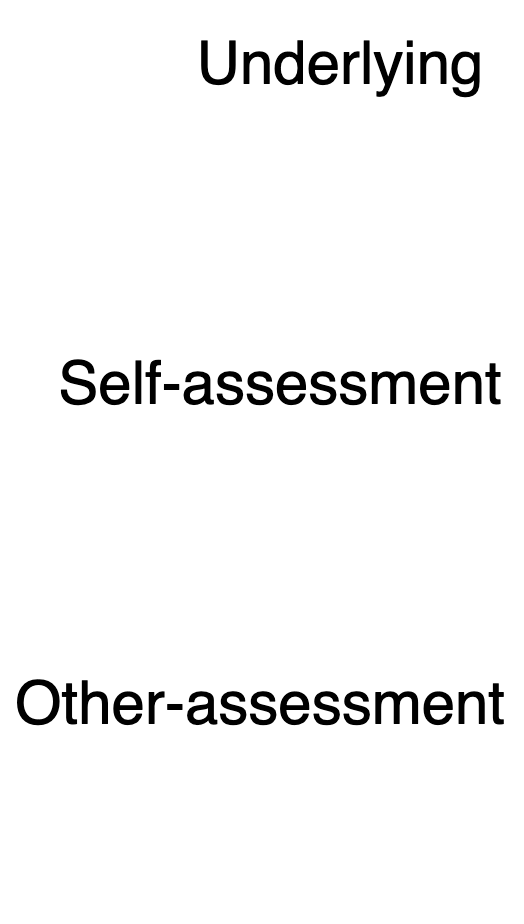}
    \end{subfigure}%
    \centering
    \begin{subfigure}{0.25\textwidth}
    \centering
        \includegraphics[width=0.8\linewidth]{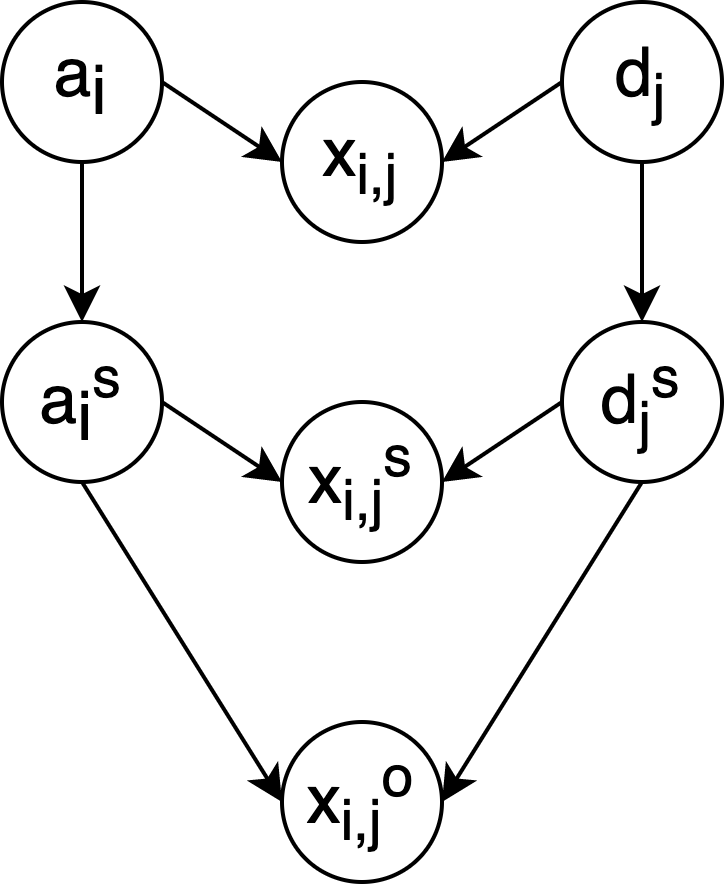}
        \caption{Undifferentiated}
        \label{fig:undiff}
    \end{subfigure}%
    \begin{subfigure}{0.25\textwidth}
    \centering
        \includegraphics[width=0.8\linewidth]{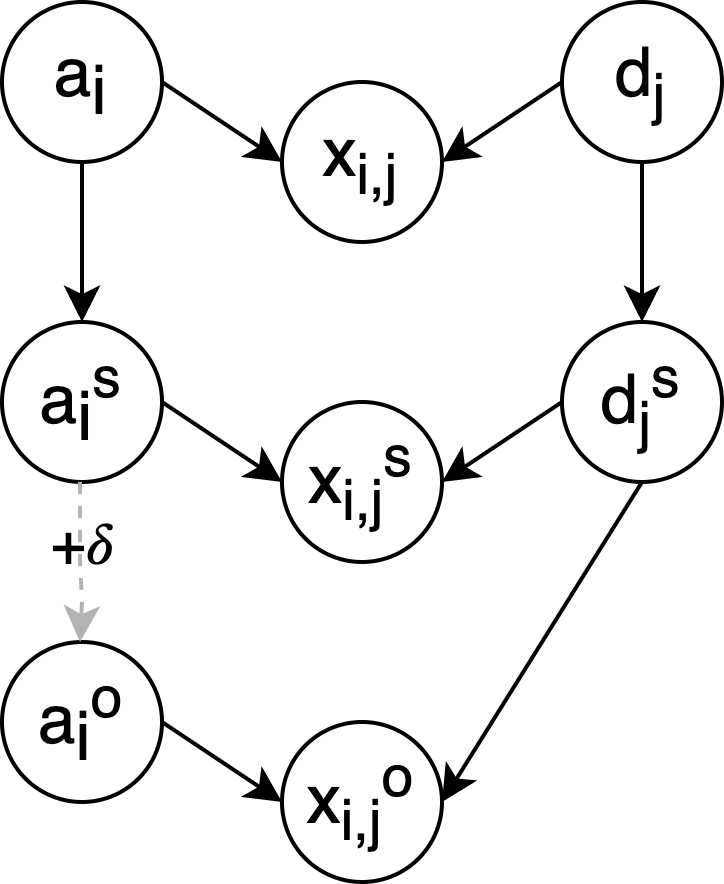}
        \caption{Differentiated by ability}
        \label{fig:diffbyability}
    \end{subfigure}%
    \begin{subfigure}{0.25\textwidth}
    \centering
        \includegraphics[width=0.8\linewidth]{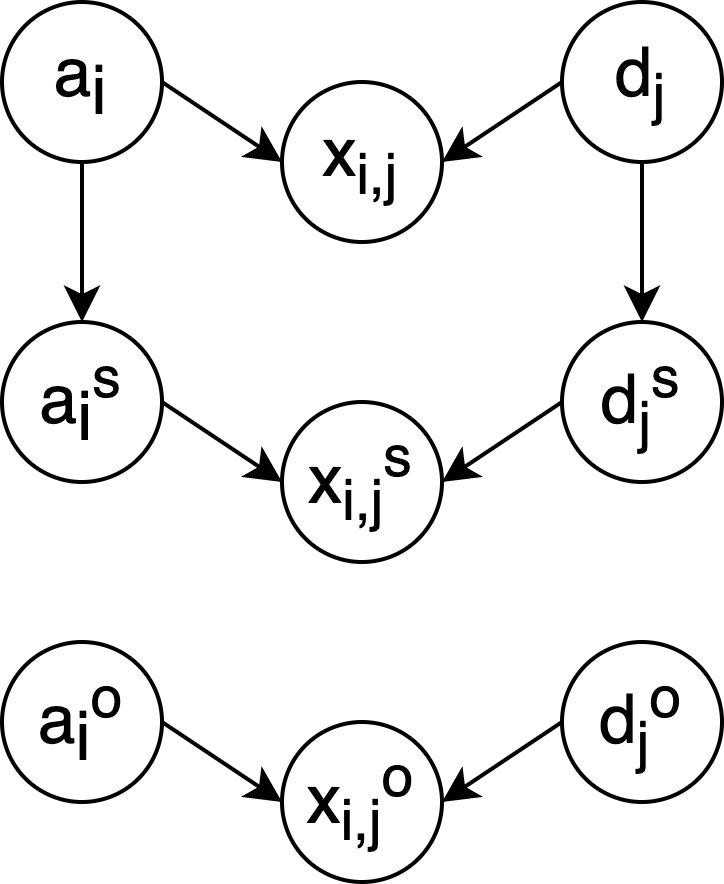}
        \caption{Fully differentiated}
        \label{fig:fullydiff}
    \end{subfigure}
    \caption{Assumed structures for the three different hierarchical models tested.}
    \label{fig:diffstructures}
\end{figure*}

We refer to the three setups as \textit{undifferentiated}, \textit{differentiated by ability}, and \textit{fully differentiated}. Figure \ref{fig:diffstructures} depicts the graphical models for each of these structures. In the \textit{undifferentiated} structure, the participant uses the same mental model to understand their own performance and the other agent's performance; the model does not allow for differentiation between the participant's own abilities and difficulties and those of the other agent. In the \textit{differentiated by ability} setup, the participant learns a difference $\delta$ between their own ability and the ability of the other agent, but problem difficulties remain undifferentiated. Finally, in the \textit{fully differentiated} structure, the person does not use their own ability or difficulties to estimate those of the other agent; the self and other mental models are independent.

We test the fit of each of these three hierarchical models to determine which matches most closely with the true relationships between parameters. Because we have evidence that these mental models are multidimensional (see Section \ref{sec:rf1}), we use a multidimensional structure for the underlying, self-assessed, and other-assessed abilities; $\delta$ is a $k$-dimensional vector capturing topic-wise ability differences.

\subsection{Results}

\subsubsection{Role of self-perception}
\label{sec:selfpresults}
For both AI agents and other people, we train three different MIRT models, one for each hypothesized hierarchical structure. To capture mental model development over time, we evaluate models based on next-round predictions, that is, we compute the log-likelihood for round $t$ using a model trained on data from rounds 1 to $t-1$. These next-round log-likelihoods, under the feedback condition, are shown in Table \ref{table:otherfb}.

\begin{table}[!ht]
\centering
\caption{Held-out next-round log-likelihoods (higher is better) for self differentiation models in the feedback condition.}
\begin{tabular}{lll}
           & Humans & AI \\
 \hline\hline
 Baseline & -2.56 & -2.56 \\
 Undifferentiated &  -2.69 & -3.36 \\
 Differentiated by Ability & \textbf{-2.00} & -2.18 \\
 Fully Differentiated & -2.13 & \textbf{-2.13} \\
\end{tabular}
\label{table:otherfb}
\end{table}

When feedback on performance is provided, the undifferentiated model has the worst fit overall, especially for AI agents---in fact, the undifferentiated model performs worse than random guessing (the baseline model). Thus, there is strong evidence that people differentiate between themselves and other agents. For perceptions of other humans, the differentiated by ability model fits the data best, suggesting that participants perceive other humans' abilities in relation to their own abilities. In contrast, the fully differentiated model best explains the perceived scores of AI agents; this provides evidence that participants' mental models of themselves were less relevant in developing mental models of AI agents (in comparison to those of other humans).


\subsubsection{Ability differential}
\label{sec:abilitydiff}
The differentiated by ability model learns a parameter $\delta$, estimating the differential between a person's self-assessed ability and their perception of the other agent's ability. These estimates, for high-accuracy other agents, are shown in Table \ref{table:deltas}. Higher values of $\delta$ correspond to higher perceived abilities of the other agent (relative to self-perceived ability). For reference, for the cities topic, $\delta=0.79$ corresponds to a 16-percentage-point increase in the latent perceived probability of a correct answer (i.e., the participant perceives the other agent's latent probability of success to be 16 percentage points higher than their own), whereas $\delta=1.72$ corresponds to a 27-percentage-point increase in that probability.\footnote{These interpretations are computed using the median values of self-perceived ability and difficulty (0.54 and -0.14, respectively).}

\begin{table}[!ht]
\centering
\caption{Latent per-topic differences between self-assessed and other-assessed ability for high-accuracy agents.}
\begin{tabular}{lll}
           & Humans & AI \\
 \hline\hline
History of Art & 0.83 & 1.63 \\
Video Games & 0.38  & 1.19 \\
Cities & 0.79  &  1.72  \\
Math &  0.75   & 1.69  \\
\end{tabular}
\label{table:deltas}
\end{table}

The variation of $\delta$s between agent types suggests substantial differences between perceptions of other humans and AI agents. Specifically, the values of $\delta$ are  larger for AI agents than for other people, despite the true actual abilities of the AI agents and other people being very similar (by design) on each topic. This aligns with the findings from Section \ref{sec:analysis}, in particular, that participants generally expect AI agents to perform at a significantly higher level than (other) humans.


\subsubsection{Relationship with self-perception}
\begin{figure*}
    \centering
    \includegraphics[width=0.45\linewidth]{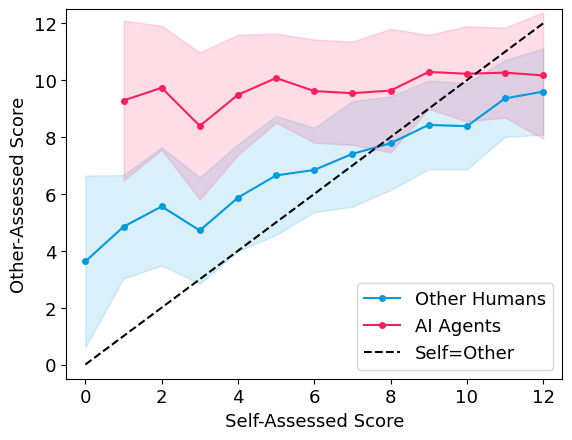}
    \caption{Relationship between self-assessed score and other-assessed score in the no feedback condition. The dotted black diagonal line represents equality between self- and other-assessment.}
    \label{fig:selfother}
\end{figure*}

In this section we investigate the overall relationship between self- and other-assessment in the absence of feedback. Note that these results, without feedback, are solely focused on each person's perception of the other agent, as a function of their self-perception; the true performance of the other agent does not play any role since no feedback is provided. Figure \ref{fig:selfother} compares self- and other-assessment scores in the no feedback condition.

Figure \ref{fig:selfother} reflects a positive correlation between self-assessment and other-assessment when evaluating other humans. In comparison, the relationship between self-assessed performance and the perceived performance of AI agents is less pronounced. When participants perceive that they have done poorly on a particular problem set, they accordingly reduce their expectations of other humans. In contrast, participants predict similar scores for AI agents across self-assessed scores. Overall, this aligns with our findings from Sections \ref{sec:selfpresults} and \ref{sec:abilitydiff}, namely, that participants expect the performance of AI agents to be more different from their own performance, compared to the performance of other humans.



We also note that for lower self-assessed scores, participants expect other agents to score higher than them, while for higher self-assessed scores, other-assessment falls slightly below self-assessment (i.e., below the diagonal). This observation aligns with existing work in cognitive science, which has found that people believe they are better than others at easier tasks, but worse than others on more difficult tasks \cite{moore2007overconfidence,dunning2011dunning}. Our experiments suggest that this finding, in particular that people believe they are worse than others on more difficult tasks, may be especially pronounced when the other is an AI agent.

\section{Discussion}
\label{sec:discussion}
First, we outline three key takeaways from our experiments and discuss how they might affect human-AI collaboration.

\paragraph{1. Over-differentiation of AI from self} We observe that mental models of an AI agent's ability are highly differentiated from self-perceived ability (e.g., Figures \ref{fig:nofeedback} and \ref{fig:withfeedback} and Table \ref{table:deltas}). On average, participants expect AI agents to perform very differently from themselves, especially in the absence of feedback. This bias could lead to under- or over-reliance on an AI agent in a team setting; additional work is needed to better understand and counteract it.

\paragraph{2. ``General intelligence'' bias} 
While we observe that people can pick up on the strengths and weaknesses of AI agents (see also \cite{bansal2019beyond,nourani2021anchoring}), in our experiments participants expect a more unified, single intelligence from an AI agent than they do another person, even after observing evidence to the contrary. In particular, the other agents in our experiments have near-identical inter-topic performance variations, but participants perceive much higher correlations between topics for AI agents (see Section \ref{sec:corrs}). This can look like a failure to recognize how well the AI agent performs in its strongest areas, and how poorly it performs in its weakest areas (see Section \ref{sec:mdtime}), potentially resulting in over- or under-use of an AI decision making aid. Again, further research is needed to determine the extent of this bias and how it might be counteracted in human-AI teams.

\paragraph{3. Incomplete development of mental models} In our experiments, participants' other-assessments did not fully converge to AI agents' true performances, even given feedback (see Figure \ref{fig:withfeedback}). This extended to the per-topic level; participants did not accurately estimate AI agents' strengths and weaknesses after 16 rounds of feedback (see Figure \ref{fig:time}). This phenomenon could serve as motivation to give the teammates of an AI agent extra information to aid in OMM development, e.g. a ``primer'' or onboarding process \cite{cai2019uncovering} or prediction explanations \cite{paleja2021utility, alipour2021improving}.


Looking ahead, we believe our modeling framework could be useful for capturing people's OMMs in the context of hybrid human-AI teams. In this paper, we investigate how a person perceives an AI agent, in terms of their different abilities, or strengths and weaknesses, and the difficulties of specific problems for the agent. Our findings and framework could help predict when a person is likely to defer to an AI agent (and thus help predict overall team performance) and identify biases that could lead to over- or under-use of the agent. 

\paragraph{Limitations} Our experiments and results are limited to a single task and setting, and involve only Amazon Mechanical Turk workers, who are not necessarily knowledgeable on the task or on AI in general. In future work, it will be important to investigate these research questions across other settings, e.g., for image classification or other tasks, and with other users, e.g., human experts who already interact with an AI agent on a regular basis.

\section{Conclusion}
\label{sec:conclusion}

In this paper, we present an experimental dataset capturing participants' mental models of themselves, other humans, and AI agents and introduce a framework for analyzing these mental models. Our findings indicate that (1) people tend to over-estimate the performance of AI agents relative to their own performance; (2) people expect the different abilities of AI agents to be highly correlated, even after observing evidence otherwise; and (3) these OMMs fail to develop completely, particularly in capturing agents' different strengths and weaknesses. We anticipate that our modeling framework, and these findings, will be useful in both understanding and improving interaction in hybrid human-AI teams.

\begin{acks}
This research was supported by NSF under awards 1900644 and 1927245, by the Irvine Initiative in AI, Law, and Society, and by the Hasso Plattner Institute (HPI) Research Center in Machine Learning and Data Science at the University of California, Irvine.
\end{acks}

\bibliographystyle{ACM-Reference-Format}
\bibliography{refs}


\begin{thebibliography}{71}


\ifx \showCODEN    \undefined \def \showCODEN     #1{\unskip}     \fi
\ifx \showDOI      \undefined \def \showDOI       #1{#1}\fi
\ifx \showISBNx    \undefined \def \showISBNx     #1{\unskip}     \fi
\ifx \showISBNxiii \undefined \def \showISBNxiii  #1{\unskip}     \fi
\ifx \showISSN     \undefined \def \showISSN      #1{\unskip}     \fi
\ifx \showLCCN     \undefined \def \showLCCN      #1{\unskip}     \fi
\ifx \shownote     \undefined \def \shownote      #1{#1}          \fi
\ifx \showarticletitle \undefined \def \showarticletitle #1{#1}   \fi
\ifx \showURL      \undefined \def \showURL       {\relax}        \fi
\providecommand\bibfield[2]{#2}
\providecommand\bibinfo[2]{#2}
\providecommand\natexlab[1]{#1}
\providecommand\showeprint[2][]{arXiv:#2}

\bibitem[Ackerman(1994)]%
        {Ackerman1994-lk}
\bibfield{author}{\bibinfo{person}{Terry~A Ackerman}.}
  \bibinfo{year}{1994}\natexlab{}.
\newblock \showarticletitle{Using multidimensional item response theory to
  understand what items and tests are measuring}.
\newblock \bibinfo{journal}{\emph{Applied Measurement in Education}}
  \bibinfo{volume}{7}, \bibinfo{number}{4} (\bibinfo{year}{1994}),
  \bibinfo{pages}{255--278}.
\newblock


\bibitem[Alipour et~al\mbox{.}(2021)]%
        {alipour2021improving}
\bibfield{author}{\bibinfo{person}{Kamran Alipour}, \bibinfo{person}{Arijit
  Ray}, \bibinfo{person}{Xiao Lin}, \bibinfo{person}{Michael Cogswell},
  \bibinfo{person}{J{\"{u}}rgen~P. Schulze}, \bibinfo{person}{Yi Yao}, {and}
  \bibinfo{person}{Giedrius~T. Burachas}.} \bibinfo{year}{2021}\natexlab{}.
\newblock \showarticletitle{Improving users' mental model with
  attention-directed counterfactual edits}.
\newblock  (\bibinfo{year}{2021}).
\newblock
\showeprint[arXiv]{2110.06863}


\bibitem[Astington and Jenkins(1995)]%
        {astington1995theory}
\bibfield{author}{\bibinfo{person}{Janet~Wilde Astington} {and}
  \bibinfo{person}{Jennifer~M Jenkins}.} \bibinfo{year}{1995}\natexlab{}.
\newblock \showarticletitle{Theory of mind development and social
  understanding}.
\newblock \bibinfo{journal}{\emph{Cognition \& Emotion}} \bibinfo{volume}{9},
  \bibinfo{number}{2-3} (\bibinfo{year}{1995}), \bibinfo{pages}{151--165}.
\newblock


\bibitem[Attenberg et~al\mbox{.}(2015)]%
        {attenberg2015beat}
\bibfield{author}{\bibinfo{person}{Joshua Attenberg}, \bibinfo{person}{Panos
  Ipeirotis}, {and} \bibinfo{person}{Foster Provost}.}
  \bibinfo{year}{2015}\natexlab{}.
\newblock \showarticletitle{Beat the machine: Challenging humans to find a
  predictive model's “unknown unknowns”}.
\newblock \bibinfo{journal}{\emph{J. Data and Information Quality}}
  \bibinfo{volume}{6}, \bibinfo{number}{1} (\bibinfo{year}{2015}),
  \bibinfo{pages}{1--17}.
\newblock


\bibitem[Bansal et~al\mbox{.}(2021a)]%
        {bansal2021teamwork}
\bibfield{author}{\bibinfo{person}{Gagan Bansal}, \bibinfo{person}{Besmira
  Nushi}, \bibinfo{person}{Ece Kamar}, \bibinfo{person}{Eric Horvitz}, {and}
  \bibinfo{person}{Daniel~S Weld}.} \bibinfo{year}{2021}\natexlab{a}.
\newblock \showarticletitle{Is the most accurate {AI} the best teammate?
  Optimizing {AI} for teamwork}. In \bibinfo{booktitle}{\emph{Proceedings of
  the AAAI Conference on Artificial Intelligence}}, Vol.~\bibinfo{volume}{35}.
  \bibinfo{pages}{11405--11414}.
\newblock


\bibitem[Bansal et~al\mbox{.}(2019a)]%
        {bansal2019beyond}
\bibfield{author}{\bibinfo{person}{Gagan Bansal}, \bibinfo{person}{Besmira
  Nushi}, \bibinfo{person}{Ece Kamar}, \bibinfo{person}{Walter~S Lasecki},
  \bibinfo{person}{Daniel~S Weld}, {and} \bibinfo{person}{Eric Horvitz}.}
  \bibinfo{year}{2019}\natexlab{a}.
\newblock \showarticletitle{Beyond accuracy: The role of mental models in
  human-{AI} team performance}. In \bibinfo{booktitle}{\emph{Proceedings of the
  AAAI Conference on Human Computation and Crowdsourcing (HCOMP 2019)}},
  Vol.~\bibinfo{volume}{7}. \bibinfo{pages}{2--11}.
\newblock


\bibitem[Bansal et~al\mbox{.}(2019b)]%
        {bansal2019updates}
\bibfield{author}{\bibinfo{person}{Gagan Bansal}, \bibinfo{person}{Besmira
  Nushi}, \bibinfo{person}{Ece Kamar}, \bibinfo{person}{Daniel~S Weld},
  \bibinfo{person}{Walter~S Lasecki}, {and} \bibinfo{person}{Eric Horvitz}.}
  \bibinfo{year}{2019}\natexlab{b}.
\newblock \showarticletitle{Updates in human-{AI} teams: Understanding and
  addressing the performance/compatibility tradeoff}. In
  \bibinfo{booktitle}{\emph{Proceedings of the AAAI Conference on Artificial
  Intelligence (AAAI 2019)}}, Vol.~\bibinfo{volume}{33}.
  \bibinfo{pages}{2429--2437}.
\newblock


\bibitem[Bansal et~al\mbox{.}(2021b)]%
        {bansal2021does}
\bibfield{author}{\bibinfo{person}{Gagan Bansal}, \bibinfo{person}{Tongshuang
  Wu}, \bibinfo{person}{Joyce Zhou}, \bibinfo{person}{Raymond Fok},
  \bibinfo{person}{Besmira Nushi}, \bibinfo{person}{Ece Kamar},
  \bibinfo{person}{Marco~Tulio Ribeiro}, {and} \bibinfo{person}{Daniel Weld}.}
  \bibinfo{year}{2021}\natexlab{b}.
\newblock \showarticletitle{Does the whole exceed its parts? The effect of {AI}
  explanations on complementary team performance}. In
  \bibinfo{booktitle}{\emph{Proceedings of the 2021 CHI Conference on Human
  Factors in Computing Systems}}. \bibinfo{pages}{1--16}.
\newblock


\bibitem[Barnes(1944)]%
        {Barnes1944-cg}
\bibfield{author}{\bibinfo{person}{Winston H~F Barnes}.}
  \bibinfo{year}{1944}\natexlab{}.
\newblock \showarticletitle{The nature of explanation}.
\newblock \bibinfo{journal}{\emph{Nature}} \bibinfo{volume}{153},
  \bibinfo{number}{3890} (\bibinfo{year}{1944}), \bibinfo{pages}{605--605}.
\newblock


\bibitem[Bichi and Talib(2018)]%
        {bichi2018item}
\bibfield{author}{\bibinfo{person}{Ado~Abdu Bichi} {and}
  \bibinfo{person}{Rohaya Talib}.} \bibinfo{year}{2018}\natexlab{}.
\newblock \showarticletitle{Item response theory: An introduction to latent
  trait models to test and item development}.
\newblock \bibinfo{journal}{\emph{International Journal of Evaluation and
  Research in Education}} \bibinfo{volume}{7}, \bibinfo{number}{2}
  (\bibinfo{year}{2018}), \bibinfo{pages}{142--151}.
\newblock


\bibitem[Bordt and Von~Luxburg(2022)]%
        {pmlr-v151-bordt22a}
\bibfield{author}{\bibinfo{person}{Sebastian Bordt} {and}
  \bibinfo{person}{Ulrike Von~Luxburg}.} \bibinfo{year}{2022}\natexlab{}.
\newblock \showarticletitle{A bandit model for human-machine decision making
  with private information and opacity}. In
  \bibinfo{booktitle}{\emph{Proceedings of the 25th International Conference on
  AI and Statistics (AI-Stats 2022)}}. \bibinfo{pages}{7300--7319}.
\newblock


\bibitem[Bos et~al\mbox{.}(2019)]%
        {bos2019mental}
\bibfield{author}{\bibinfo{person}{Nathan Bos}, \bibinfo{person}{Kimberly
  Glasgow}, \bibinfo{person}{John Gersh}, \bibinfo{person}{Isaiah Harbison},
  {and} \bibinfo{person}{Celeste Lyn~Paul}.} \bibinfo{year}{2019}\natexlab{}.
\newblock \showarticletitle{Mental models of {AI}-based systems: User
  predictions and explanations of image classification results}. In
  \bibinfo{booktitle}{\emph{Proceedings of the Human Factors and Ergonomics
  Society Annual Meeting}}, Vol.~\bibinfo{volume}{63}.
  \bibinfo{pages}{184--188}.
\newblock


\bibitem[Boyd-Graber and B{\"o}rschinger(2020)]%
        {graber2019trivia}
\bibfield{author}{\bibinfo{person}{Jordan Boyd-Graber} {and}
  \bibinfo{person}{Benjamin B{\"o}rschinger}.} \bibinfo{year}{2020}\natexlab{}.
\newblock \showarticletitle{What question answering can learn from trivia
  nerds}. In \bibinfo{booktitle}{\emph{Proceedings of the 58th Annual Meeting
  of the Association for Computational Linguistics}}.
  \bibinfo{pages}{7422–7435}.
\newblock


\bibitem[Buehler and Weisswange(2020)]%
        {buehler2020theory}
\bibfield{author}{\bibinfo{person}{Moritz~C. Buehler} {and}
  \bibinfo{person}{Thomas~H. Weisswange}.} \bibinfo{year}{2020}\natexlab{}.
\newblock \showarticletitle{Theory of mind based communication for human agent
  cooperation}. In \bibinfo{booktitle}{\emph{2020 IEEE International Conference
  on Human-Machine Systems (ICHMS)}}. \bibinfo{pages}{1--6}.
\newblock


\bibitem[Cai et~al\mbox{.}(2019)]%
        {cai2019uncovering}
\bibfield{author}{\bibinfo{person}{Carrie~J. Cai}, \bibinfo{person}{Samantha
  Winter}, \bibinfo{person}{David Steiner}, \bibinfo{person}{Lauren Wilcox},
  {and} \bibinfo{person}{Michael Terry}.} \bibinfo{year}{2019}\natexlab{}.
\newblock \showarticletitle{"Hello {AI}": Uncovering the onboarding needs of
  medical practitioners for human-{AI} collaborative decision-making}.
\newblock \bibinfo{journal}{\emph{Proceedings of the ACM on Human-Computer
  Interaction}}  \bibinfo{volume}{3}, Article \bibinfo{articleno}{104}
  (\bibinfo{year}{2019}), \bibinfo{numpages}{24}~pages.
\newblock


\bibitem[Castelo et~al\mbox{.}(2019)]%
        {castelo2019task}
\bibfield{author}{\bibinfo{person}{Noah Castelo}, \bibinfo{person}{Maarten~W
  Bos}, {and} \bibinfo{person}{Donald~R Lehmann}.}
  \bibinfo{year}{2019}\natexlab{}.
\newblock \showarticletitle{Task-dependent algorithm aversion}.
\newblock \bibinfo{journal}{\emph{Journal of Marketing Research}}
  \bibinfo{volume}{56}, \bibinfo{number}{5} (\bibinfo{year}{2019}),
  \bibinfo{pages}{809--825}.
\newblock


\bibitem[Chandler et~al\mbox{.}(2022)]%
        {chandler2022applicability}
\bibfield{author}{\bibinfo{person}{Chelsea Chandler}, \bibinfo{person}{Peter~W
  Foltz}, {and} \bibinfo{person}{Brita Elvevåg}.}
  \bibinfo{year}{2022}\natexlab{}.
\newblock \showarticletitle{{improving the applicability of {AI} for
  psychiatric applications through human-in-the-loop methodologies}}.
\newblock \bibinfo{journal}{\emph{Schizophrenia Bulletin}}
  \bibinfo{volume}{48}, \bibinfo{number}{5} (\bibinfo{year}{2022}),
  \bibinfo{pages}{949--957}.
\newblock


\bibitem[Chen et~al\mbox{.}(2023)]%
        {chen2023understanding}
\bibfield{author}{\bibinfo{person}{Valerie Chen}, \bibinfo{person}{Q.~Vera
  Liao}, \bibinfo{person}{Jennifer~Wortman Vaughan}, {and}
  \bibinfo{person}{Gagan Bansal}.} \bibinfo{year}{2023}\natexlab{}.
\newblock \showarticletitle{Understanding the Role of Human Intuition on
  Reliance in Human-AI Decision-Making with Explanations}.
\newblock \bibinfo{journal}{\emph{arXiv preprint arXiv:2301.07255}}
  (\bibinfo{year}{2023}).
\newblock


\bibitem[Cheng et~al\mbox{.}(2022)]%
        {cheng2022welfare}
\bibfield{author}{\bibinfo{person}{Hao-Fei Cheng}, \bibinfo{person}{Logan
  Stapleton}, \bibinfo{person}{Anna Kawakami}, \bibinfo{person}{Venkatesh
  Sivaraman}, \bibinfo{person}{Yanghuidi Cheng}, \bibinfo{person}{Diana Qing},
  \bibinfo{person}{Adam Perer}, \bibinfo{person}{Kenneth Holstein},
  \bibinfo{person}{Zhiwei~Steven Wu}, {and} \bibinfo{person}{Haiyi Zhu}.}
  \bibinfo{year}{2022}\natexlab{}.
\newblock \showarticletitle{How child welfare workers reduce racial disparities
  in algorithmic decisions}. In \bibinfo{booktitle}{\emph{CHI '22: Proceedings
  of the 2022 CHI Conference on Human Factors in Computing Systems}}.
  \bibinfo{pages}{1--22}.
\newblock


\bibitem[De-Arteaga et~al\mbox{.}(2020)]%
        {de2020case}
\bibfield{author}{\bibinfo{person}{Maria De-Arteaga}, \bibinfo{person}{Riccardo
  Fogliato}, {and} \bibinfo{person}{Alexandra Chouldechova}.}
  \bibinfo{year}{2020}\natexlab{}.
\newblock \showarticletitle{A case for humans-in-the-loop: Decisions in the
  presence of erroneous algorithmic scores}. In
  \bibinfo{booktitle}{\emph{Proceedings of the 2020 CHI Conference on Human
  Factors in Computing Systems}}. \bibinfo{pages}{1--12}.
\newblock


\bibitem[d'Eon et~al\mbox{.}(2022)]%
        {d2022spotlight}
\bibfield{author}{\bibinfo{person}{Greg d'Eon}, \bibinfo{person}{Jason d'Eon},
  \bibinfo{person}{James~R Wright}, {and} \bibinfo{person}{Kevin
  Leyton-Brown}.} \bibinfo{year}{2022}\natexlab{}.
\newblock \showarticletitle{The Spotlight: A general method for discovering
  systematic errors in deep learning models}. In
  \bibinfo{booktitle}{\emph{Proceedings of the 2022 ACM Conference on Fairness,
  Accountability, and Transparency}}. \bibinfo{pages}{1962--1981}.
\newblock


\bibitem[Donahue et~al\mbox{.}(2022)]%
        {donahue2022unfairness}
\bibfield{author}{\bibinfo{person}{Kate Donahue}, \bibinfo{person}{Alexandra
  Chouldechova}, {and} \bibinfo{person}{Krishnaram Kenthapadi}.}
  \bibinfo{year}{2022}\natexlab{}.
\newblock \showarticletitle{Human-algorithm collaboration: Achieving
  complementarity and avoiding unfairness}. In
  \bibinfo{booktitle}{\emph{Proceedings of the 2022 ACM Conference on Fairness,
  Accountability, and Transparency}}. \bibinfo{pages}{1639–1656}.
\newblock


\bibitem[Druce et~al\mbox{.}(2021)]%
        {druce2021brittle}
\bibfield{author}{\bibinfo{person}{Jeff Druce}, \bibinfo{person}{James
  Niehaus}, \bibinfo{person}{Vanessa Moody}, \bibinfo{person}{David~D. Jensen},
  {and} \bibinfo{person}{Michael~L. Littman}.} \bibinfo{year}{2021}\natexlab{}.
\newblock \showarticletitle{Brittle {AI}, causal confusion, and bad mental
  models: challenges and successes in the {XAI} program}.
\newblock \bibinfo{journal}{\emph{CoRR}}  \bibinfo{volume}{abs/2106.05506}
  (\bibinfo{year}{2021}).
\newblock
\showeprint[arXiv]{2106.05506}


\bibitem[Dunlosky and Metcalfe(2008)]%
        {dunlosky2008metacognition}
\bibfield{author}{\bibinfo{person}{John Dunlosky} {and} \bibinfo{person}{Janet
  Metcalfe}.} \bibinfo{year}{2008}\natexlab{}.
\newblock \bibinfo{booktitle}{\emph{Metacognition}}.
\newblock \bibinfo{publisher}{Sage Publications}.
\newblock


\bibitem[Dunning(2011)]%
        {dunning2011dunning}
\bibfield{author}{\bibinfo{person}{David Dunning}.}
  \bibinfo{year}{2011}\natexlab{}.
\newblock \showarticletitle{The {D}unning--{K}ruger effect: On being ignorant
  of one's own ignorance}.
\newblock In \bibinfo{booktitle}{\emph{Advances in Experimental Social
  Psychology}}. Vol.~\bibinfo{volume}{44}. \bibinfo{pages}{247--296}.
\newblock


\bibitem[Fox(2010)]%
        {fox2010bayesian}
\bibfield{author}{\bibinfo{person}{Jean-Paul Fox}.}
  \bibinfo{year}{2010}\natexlab{}.
\newblock \bibinfo{booktitle}{\emph{Bayesian Item Response Modeling: Theory and
  Applications}}.
\newblock \bibinfo{publisher}{Springer, New York}.
\newblock


\bibitem[Frith and Frith(2005)]%
        {frith2005theory}
\bibfield{author}{\bibinfo{person}{Chris Frith} {and} \bibinfo{person}{Uta
  Frith}.} \bibinfo{year}{2005}\natexlab{}.
\newblock \showarticletitle{Theory of mind}.
\newblock \bibinfo{journal}{\emph{Current Biology}} \bibinfo{volume}{15},
  \bibinfo{number}{17} (\bibinfo{year}{2005}), \bibinfo{pages}{R644--R645}.
\newblock


\bibitem[Gero et~al\mbox{.}(2020)]%
        {gero2020mental}
\bibfield{author}{\bibinfo{person}{Katy~Ilonka Gero}, \bibinfo{person}{Zahra
  Ashktorab}, \bibinfo{person}{Casey Dugan}, \bibinfo{person}{Qian Pan},
  \bibinfo{person}{James Johnson}, \bibinfo{person}{Werner Geyer},
  \bibinfo{person}{Maria Ruiz}, \bibinfo{person}{Sarah Miller},
  \bibinfo{person}{David~R Millen}, \bibinfo{person}{Murray Campbell},
  {et~al\mbox{.}}} \bibinfo{year}{2020}\natexlab{}.
\newblock \showarticletitle{Mental models of AI agents in a cooperative game
  setting}. In \bibinfo{booktitle}{\emph{Proceedings of the 2020 CHI Conference
  on Human Factors in Computing Systems}}. \bibinfo{pages}{1--12}.
\newblock


\bibitem[Hartig and Höhler(2009)]%
        {HARTIG200957}
\bibfield{author}{\bibinfo{person}{Johannes Hartig} {and} \bibinfo{person}{Jana
  Höhler}.} \bibinfo{year}{2009}\natexlab{}.
\newblock \showarticletitle{Multidimensional {IRT} models for the assessment of
  competencies}.
\newblock \bibinfo{journal}{\emph{Studies in Educational Evaluation}}
  \bibinfo{volume}{35}, \bibinfo{number}{2} (\bibinfo{year}{2009}),
  \bibinfo{pages}{57--63}.
\newblock


\bibitem[Hemmer et~al\mbox{.}(2021)]%
        {hemmer2021human}
\bibfield{author}{\bibinfo{person}{Patrick Hemmer}, \bibinfo{person}{Max
  Schemmer}, \bibinfo{person}{Michael V{\"o}ssing}, {and}
  \bibinfo{person}{Niklas K{\"u}hl}.} \bibinfo{year}{2021}\natexlab{}.
\newblock \showarticletitle{Human-{AI} complementarity in hybrid intelligence
  systems: A structured literature review}. In
  \bibinfo{booktitle}{\emph{Proceedings of the Twenty-fifth Pacific Asia
  Conference on Information Systems,}}. \bibinfo{pages}{1--14}.
\newblock


\bibitem[Holstein and Aleven(2022)]%
        {holstein2021k12}
\bibfield{author}{\bibinfo{person}{Kenneth Holstein} {and}
  \bibinfo{person}{Vincent Aleven}.} \bibinfo{year}{2022}\natexlab{}.
\newblock \showarticletitle{Designing for human-{AI} complementarity in {K-12}
  education}.
\newblock \bibinfo{journal}{\emph{AI Magazine}} \bibinfo{volume}{43},
  \bibinfo{number}{2} (\bibinfo{year}{2022}), \bibinfo{pages}{239--248}.
\newblock


\bibitem[Humphreys(1979)]%
        {HUMPHREYS1979105}
\bibfield{author}{\bibinfo{person}{G. Humphreys, Lloyd}.}
  \bibinfo{year}{1979}\natexlab{}.
\newblock \showarticletitle{The construct of general intelligence}.
\newblock \bibinfo{journal}{\emph{Intelligence}} \bibinfo{volume}{3},
  \bibinfo{number}{2} (\bibinfo{year}{1979}), \bibinfo{pages}{105--120}.
\newblock


\bibitem[Kamar(2016)]%
        {kamar2016directions}
\bibfield{author}{\bibinfo{person}{Ece Kamar}.}
  \bibinfo{year}{2016}\natexlab{}.
\newblock \showarticletitle{Directions in hybrid iIntelligence: Complementing
  {AI} systems with human intelligence.}. In
  \bibinfo{booktitle}{\emph{Proceedings of the International Joint Conference
  on AI (IJCAI 2016)}}. \bibinfo{pages}{4070--4073}.
\newblock


\bibitem[Kamar et~al\mbox{.}(2012)]%
        {kamar2012combining}
\bibfield{author}{\bibinfo{person}{Ece Kamar}, \bibinfo{person}{Severin
  Hacker}, {and} \bibinfo{person}{Eric Horvitz}.}
  \bibinfo{year}{2012}\natexlab{}.
\newblock \showarticletitle{Combining human and machine intelligence in
  large-scale crowdsourcing.}. In \bibinfo{booktitle}{\emph{Proceedings of the
  AAMAS Conference}}, Vol.~\bibinfo{volume}{12}. \bibinfo{pages}{467--474}.
\newblock


\bibitem[Khashabi et~al\mbox{.}(2020)]%
        {khashabi2020unifiedqa}
\bibfield{author}{\bibinfo{person}{Daniel Khashabi}, \bibinfo{person}{Sewon
  Min}, \bibinfo{person}{Tushar Khot}, \bibinfo{person}{Ashish Sabharwal},
  \bibinfo{person}{Oyvind Tafjord}, \bibinfo{person}{Peter Clark}, {and}
  \bibinfo{person}{Hannaneh Hajishirzi}.} \bibinfo{year}{2020}\natexlab{}.
\newblock \showarticletitle{UnifiedQA: Crossing format boundaries with a single
  {QA} system}.
\newblock \bibinfo{journal}{\emph{arXiv preprint arXiv:2005.00700}}
  (\bibinfo{year}{2020}).
\newblock


\bibitem[Kojima et~al\mbox{.}(2022)]%
        {kojima2022large}
\bibfield{author}{\bibinfo{person}{Takeshi Kojima},
  \bibinfo{person}{Shixiang~Shane Gu}, \bibinfo{person}{Machel Reid},
  \bibinfo{person}{Yutaka Matsuo}, {and} \bibinfo{person}{Yusuke Iwasawa}.}
  \bibinfo{year}{2022}\natexlab{}.
\newblock \showarticletitle{Large language models are zero-shot reasoners}. In
  \bibinfo{booktitle}{\emph{Advances in Neural Information Processing
  Systems}}. \bibinfo{pages}{22199--22213}.
\newblock


\bibitem[Kulesza et~al\mbox{.}(2012)]%
        {kulesza2012tell}
\bibfield{author}{\bibinfo{person}{Todd Kulesza}, \bibinfo{person}{Simone
  Stumpf}, \bibinfo{person}{Margaret Burnett}, {and} \bibinfo{person}{Irwin
  Kwan}.} \bibinfo{year}{2012}\natexlab{}.
\newblock \showarticletitle{Tell me more? The effects of mental model soundness
  on personalizing an intelligent agent}. In
  \bibinfo{booktitle}{\emph{Proceedings of the SIGCHI Conference on Human
  Factors in Computing Systems}}. \bibinfo{pages}{1–10}.
\newblock


\bibitem[Kumar et~al\mbox{.}(2023)]%
        {kumar2023differentiating}
\bibfield{author}{\bibinfo{person}{Aakriti Kumar}, \bibinfo{person}{Padhraic
  Smyth}, {and} \bibinfo{person}{Mark Steyvers}.}
  \bibinfo{year}{2023}\natexlab{}.
\newblock \showarticletitle{Differentiating mental models of self and others: a
  hierarchical framework for knowledge assessment}.
\newblock \bibinfo{journal}{\emph{PsyArXiv}} (\bibinfo{year}{2023}).
\newblock


\bibitem[La~Barbera et~al\mbox{.}(2022)]%
        {la2022hybrid}
\bibfield{author}{\bibinfo{person}{David La~Barbera}, \bibinfo{person}{Kevin
  Roitero}, {and} \bibinfo{person}{Stefano Mizzaro}.}
  \bibinfo{year}{2022}\natexlab{}.
\newblock \showarticletitle{A hybrid human-in-the-loop framework for fact
  checking}. In \bibinfo{booktitle}{\emph{Proceedings of the Sixth Workshop on
  Natural Language for Artificial Intelligence (NL4AI 2022)}}.
\newblock


\bibitem[Lai et~al\mbox{.}(2021)]%
        {lai2021science}
\bibfield{author}{\bibinfo{person}{Vivian Lai}, \bibinfo{person}{Chacha Chen},
  \bibinfo{person}{Q.~Vera Liao}, \bibinfo{person}{Alison Smith{-}Renner},
  {and} \bibinfo{person}{Chenhao Tan}.} \bibinfo{year}{2021}\natexlab{}.
\newblock \showarticletitle{Towards a science of human-{AI} decision making:
  {A} survey of empirical studies}.
\newblock \bibinfo{journal}{\emph{CoRR}}  \bibinfo{volume}{abs/2112.11471}
  (\bibinfo{year}{2021}).
\newblock
\urldef\tempurl%
\url{https://arxiv.org/abs/2112.11471}
\showURL{%
\tempurl}


\bibitem[lai Lee et~al\mbox{.}(2005)]%
        {lee2005robots}
\bibfield{author}{\bibinfo{person}{Sau lai Lee}, \bibinfo{person}{Ivy~Yee man
  Lau}, \bibinfo{person}{S. Kiesler}, {and} \bibinfo{person}{Chi-Yue Chiu}.}
  \bibinfo{year}{2005}\natexlab{}.
\newblock \showarticletitle{Human mental models of humanoid robots}. In
  \bibinfo{booktitle}{\emph{Proceedings of the 2005 IEEE International
  Conference on Robotics and Automation}}. \bibinfo{pages}{2767--2772}.
\newblock


\bibitem[Liang et~al\mbox{.}(2022)]%
        {liang2022adapting}
\bibfield{author}{\bibinfo{person}{Garston Liang}, \bibinfo{person}{Jennifer~F
  Sloane}, \bibinfo{person}{Christopher Donkin}, {and} \bibinfo{person}{Ben~R
  Newell}.} \bibinfo{year}{2022}\natexlab{}.
\newblock \showarticletitle{Adapting to the algorithm: How accuracy comparisons
  promote the use of a decision aid}.
\newblock \bibinfo{journal}{\emph{Cognitive Research: Principles and
  Implications}} \bibinfo{volume}{7}, \bibinfo{number}{1}
  (\bibinfo{year}{2022}), \bibinfo{pages}{14}.
\newblock


\bibitem[Livingston(2003)]%
        {livingston2003metacognition}
\bibfield{author}{\bibinfo{person}{Jennifer~A Livingston}.}
  \bibinfo{year}{2003}\natexlab{}.
\newblock \bibinfo{title}{Metacognition: An Overview}.
\newblock
\newblock


\bibitem[Logg(2017)]%
        {logg2017theory}
\bibfield{author}{\bibinfo{person}{Jennifer~Marie Logg}.}
  \bibinfo{year}{2017}\natexlab{}.
\newblock \showarticletitle{Theory of machine: When do people rely on
  algorithms?}
\newblock \bibinfo{journal}{\emph{Harvard Business School working paper
  series\# 17-086}} (\bibinfo{year}{2017}).
\newblock


\bibitem[Logg(2022)]%
        {logg2022psychology}
\bibfield{author}{\bibinfo{person}{Jennifer~M Logg}.}
  \bibinfo{year}{2022}\natexlab{}.
\newblock \showarticletitle{The psychology of big data: Developing a “theory
  of machine” to examine perceptions of algorithms}.
\newblock In \bibinfo{booktitle}{\emph{The Psychology of Technology: Social
  Science Research in the Age of Big Data}},
  \bibfield{editor}{\bibinfo{person}{Sandra Matz}} (Ed.).
  \bibinfo{publisher}{American Psychological Association},
  \bibinfo{pages}{349--378}.
\newblock


\bibitem[Luo and Al-Harbi(2017)]%
        {luo2017performances}
\bibfield{author}{\bibinfo{person}{Yong Luo} {and} \bibinfo{person}{Khaleel
  Al-Harbi}.} \bibinfo{year}{2017}\natexlab{}.
\newblock \showarticletitle{Performances of LOO and WAIC as IRT model selection
  methods}.
\newblock \bibinfo{journal}{\emph{Psychological Test and Assessment Modeling}}
  \bibinfo{volume}{59}, \bibinfo{number}{2} (\bibinfo{year}{2017}),
  \bibinfo{pages}{183}.
\newblock


\bibitem[Mathieu et~al\mbox{.}(2000)]%
        {matheieu2000shared}
\bibfield{author}{\bibinfo{person}{John Mathieu}, \bibinfo{person}{Tonia
  Heffner}, \bibinfo{person}{Gerald Goodwin}, \bibinfo{person}{Eduardo Salas},
  {and} \bibinfo{person}{Janis Cannon-Bowers}.}
  \bibinfo{year}{2000}\natexlab{}.
\newblock \showarticletitle{The influence of shared mental models on team
  process and performance}.
\newblock \bibinfo{journal}{\emph{Journal of Applied Psychology}}
  \bibinfo{volume}{85} (\bibinfo{date}{04} \bibinfo{year}{2000}),
  \bibinfo{pages}{273--283}.
\newblock


\bibitem[Merry et~al\mbox{.}(2021)]%
        {merry2021mental}
\bibfield{author}{\bibinfo{person}{Michael Merry}, \bibinfo{person}{Pat
  Riddle}, {and} \bibinfo{person}{Jim Warren}.}
  \bibinfo{year}{2021}\natexlab{}.
\newblock \showarticletitle{A mental models approach for defining explainable
  artificial intelligence}.
\newblock \bibinfo{journal}{\emph{BMC Medical Informatics and Decision Making}}
  \bibinfo{volume}{21}, \bibinfo{number}{1} (\bibinfo{year}{2021}),
  \bibinfo{pages}{1--12}.
\newblock


\bibitem[Miller(2019)]%
        {miller2019explanation}
\bibfield{author}{\bibinfo{person}{Tim Miller}.}
  \bibinfo{year}{2019}\natexlab{}.
\newblock \showarticletitle{Explanation in artificial intelligence: Insights
  from the social sciences}.
\newblock \bibinfo{journal}{\emph{Artificial intelligence}}
  \bibinfo{volume}{267} (\bibinfo{year}{2019}), \bibinfo{pages}{1--38}.
\newblock


\bibitem[Moore and Cain(2007)]%
        {moore2007overconfidence}
\bibfield{author}{\bibinfo{person}{Don~A Moore} {and}
  \bibinfo{person}{Daylian~M Cain}.} \bibinfo{year}{2007}\natexlab{}.
\newblock \showarticletitle{Overconfidence and underconfidence: When and why
  people underestimate (and overestimate) the competition}.
\newblock \bibinfo{journal}{\emph{Organizational Behavior and Human Decision
  Processes}} \bibinfo{volume}{103}, \bibinfo{number}{2}
  (\bibinfo{year}{2007}), \bibinfo{pages}{197--213}.
\newblock


\bibitem[Nourani et~al\mbox{.}(2021)]%
        {nourani2021anchoring}
\bibfield{author}{\bibinfo{person}{Mahsan Nourani}, \bibinfo{person}{Chiradeep
  Roy}, \bibinfo{person}{Jeremy~E Block}, \bibinfo{person}{Donald~R Honeycutt},
  \bibinfo{person}{Tahrima Rahman}, \bibinfo{person}{Eric Ragan}, {and}
  \bibinfo{person}{Vibhav Gogate}.} \bibinfo{year}{2021}\natexlab{}.
\newblock \showarticletitle{Anchoring bias affects mental model formation and
  user reliance in explainable {AI} systems}. In
  \bibinfo{booktitle}{\emph{Proceedings of the 26th International Conference on
  Intelligent User Interfaces}}. \bibinfo{pages}{340–350}.
\newblock


\bibitem[Paleja et~al\mbox{.}(2021)]%
        {paleja2021utility}
\bibfield{author}{\bibinfo{person}{Rohan Paleja}, \bibinfo{person}{Muyleng
  Ghuy}, \bibinfo{person}{Nadun Ranawaka~Arachchige}, \bibinfo{person}{Reed
  Jensen}, {and} \bibinfo{person}{Matthew Gombolay}.}
  \bibinfo{year}{2021}\natexlab{}.
\newblock \showarticletitle{The utility of explainable {AI} in ad hoc
  human-machine teaming}. In \bibinfo{booktitle}{\emph{Advances in Neural
  Information Processing Systems}}, Vol.~\bibinfo{volume}{34}.
  \bibinfo{pages}{610--623}.
\newblock


\bibitem[Rastogi et~al\mbox{.}(2022)]%
        {rastogi2022unifying}
\bibfield{author}{\bibinfo{person}{Charvi Rastogi}, \bibinfo{person}{Liu Leqi},
  \bibinfo{person}{Kenneth Holstein}, {and} \bibinfo{person}{Hoda Heidari}.}
  \bibinfo{year}{2022}\natexlab{}.
\newblock \showarticletitle{A unifying framework for combining complementary
  strengths of humans and {ML} toward better predictive decision-making}.
\newblock \bibinfo{journal}{\emph{arXiv preprint arXiv:2204.10806}}
  (\bibinfo{year}{2022}).
\newblock


\bibitem[Reckase(1997)]%
        {reckase1997future}
\bibfield{author}{\bibinfo{person}{Mark~D. Reckase}.}
  \bibinfo{year}{1997}\natexlab{}.
\newblock \showarticletitle{The past and future of multidimensional item
  response theory}.
\newblock \bibinfo{journal}{\emph{Applied Psychological Measurement}}
  \bibinfo{volume}{21}, \bibinfo{number}{1} (\bibinfo{year}{1997}),
  \bibinfo{pages}{25--36}.
\newblock


\bibitem[Schelble et~al\mbox{.}(2022)]%
        {schelble2022shared}
\bibfield{author}{\bibinfo{person}{Beau~G. Schelble},
  \bibinfo{person}{Christopher Flathmann}, \bibinfo{person}{Nathan~J. McNeese},
  \bibinfo{person}{Guo Freeman}, {and} \bibinfo{person}{Rohit Mallick}.}
  \bibinfo{year}{2022}\natexlab{}.
\newblock \showarticletitle{Let's think together! Assessing shared mental
  models, performance, and trust in human-agent teams}.
\newblock \bibinfo{journal}{\emph{Proceedings of the ACM on Human-Computer
  Interaction}}  \bibinfo{volume}{6}, Article \bibinfo{articleno}{13}
  (\bibinfo{year}{2022}), \bibinfo{numpages}{29}~pages.
\newblock


\bibitem[Scheutz et~al\mbox{.}(2017)]%
        {scheutz2017framework}
\bibfield{author}{\bibinfo{person}{Matthias Scheutz}, \bibinfo{person}{Scott~A
  DeLoach}, {and} \bibinfo{person}{Julie~A Adams}.}
  \bibinfo{year}{2017}\natexlab{}.
\newblock \showarticletitle{A framework for developing and using shared mental
  models in human-agent teams}.
\newblock \bibinfo{journal}{\emph{Journal of Cognitive Engineering and Decision
  Making}} \bibinfo{volume}{11}, \bibinfo{number}{3} (\bibinfo{year}{2017}),
  \bibinfo{pages}{203--224}.
\newblock


\bibitem[Sheng and Wikle(2007)]%
        {Sheng2007-mi}
\bibfield{author}{\bibinfo{person}{Yanyan Sheng} {and}
  \bibinfo{person}{Christopher~K Wikle}.} \bibinfo{year}{2007}\natexlab{}.
\newblock \showarticletitle{Comparing multiunidimensional and unidimensional
  item response theory models}.
\newblock \bibinfo{journal}{\emph{Educ. Psychol. Meas.}} \bibinfo{volume}{67},
  \bibinfo{number}{6} (\bibinfo{year}{2007}), \bibinfo{pages}{899--919}.
\newblock


\bibitem[Sidera et~al\mbox{.}(2018)]%
        {sidera2018theory}
\bibfield{author}{\bibinfo{person}{Francesc Sidera}, \bibinfo{person}{Georgina
  Perpi{\~n}{\`a}}, \bibinfo{person}{J{\`e}ssica Serrano}, {and}
  \bibinfo{person}{Carles Rostan}.} \bibinfo{year}{2018}\natexlab{}.
\newblock \showarticletitle{Why is theory of mind important for referential
  communication?}
\newblock \bibinfo{journal}{\emph{Current Psychology}}  \bibinfo{volume}{37}
  (\bibinfo{year}{2018}), \bibinfo{pages}{82--97}.
\newblock


\bibitem[Smyth et~al\mbox{.}(1994)]%
        {Smyth1994-qo}
\bibfield{author}{\bibinfo{person}{Mary~M Smyth}, \bibinfo{person}{Alan~F
  Collins}, \bibinfo{person}{Peter~E Morris}, {and} \bibinfo{person}{Philip
  Levy}.} \bibinfo{year}{1994}\natexlab{}.
\newblock \bibinfo{booktitle}{\emph{Cognition in Action}
  (\bibinfo{edition}{2nd} ed.)}.
\newblock \bibinfo{publisher}{Lawrence Erlbaum Associates}.
\newblock


\bibitem[Spearman(1904)]%
        {spearman1904general}
\bibfield{author}{\bibinfo{person}{C. Spearman}.}
  \bibinfo{year}{1904}\natexlab{}.
\newblock \showarticletitle{"General intelligence," objectively determined and
  measured}.
\newblock \bibinfo{journal}{\emph{The American Journal of Psychology}}
  \bibinfo{volume}{15}, \bibinfo{number}{2} (\bibinfo{year}{1904}),
  \bibinfo{pages}{201--292}.
\newblock


\bibitem[Steyvers and Kumar(2022)]%
        {steyvers_kumar_2022}
\bibfield{author}{\bibinfo{person}{Mark Steyvers} {and}
  \bibinfo{person}{Aakriti Kumar}.} \bibinfo{year}{2022}\natexlab{}.
\newblock \showarticletitle{Three challenges for {AI}-assisted
  decision-making}.
\newblock \bibinfo{journal}{\emph{PsyArXiv}} (\bibinfo{year}{2022}).
\newblock
\urldef\tempurl%
\url{https://doi.org/10.31234/osf.io/gctv6}
\showDOI{\tempurl}


\bibitem[Steyvers et~al\mbox{.}(2022)]%
        {steyvers2022bayesian}
\bibfield{author}{\bibinfo{person}{Mark Steyvers}, \bibinfo{person}{Heliodoro
  Tejeda}, \bibinfo{person}{Gavin Kerrigan}, {and} \bibinfo{person}{Padhraic
  Smyth}.} \bibinfo{year}{2022}\natexlab{}.
\newblock \showarticletitle{Bayesian modeling of human–{AI} complementarity}.
\newblock \bibinfo{journal}{\emph{Proceedings of the National Academy of
  Sciences}} \bibinfo{volume}{119}, \bibinfo{number}{11}
  (\bibinfo{year}{2022}), \bibinfo{pages}{e2111547119}.
\newblock


\bibitem[Thomas(2019)]%
        {thomas2019advances}
\bibfield{author}{\bibinfo{person}{Michael~L Thomas}.}
  \bibinfo{year}{2019}\natexlab{}.
\newblock \showarticletitle{Advances in applications of item response theory to
  clinical assessment}.
\newblock \bibinfo{journal}{\emph{Psychological Assessment}}
  \bibinfo{volume}{31}, \bibinfo{number}{12} (\bibinfo{year}{2019}),
  \bibinfo{pages}{1442--1455}.
\newblock


\bibitem[Tschandl et~al\mbox{.}(2020)]%
        {Tschandl2020yv}
\bibfield{author}{\bibinfo{person}{Philipp Tschandl},
  \bibinfo{person}{Christoph Rinner}, \bibinfo{person}{Zoe Apalla},
  \bibinfo{person}{Giuseppe Argenziano}, \bibinfo{person}{Noel Codella},
  \bibinfo{person}{Allan Halpern}, \bibinfo{person}{Monika Janda},
  \bibinfo{person}{Aimilios Lallas}, \bibinfo{person}{Caterina Longo},
  \bibinfo{person}{Josep Malvehy}, \bibinfo{person}{John Paoli},
  \bibinfo{person}{Susana Puig}, \bibinfo{person}{Cliff Rosendahl},
  \bibinfo{person}{H~Peter Soyer}, \bibinfo{person}{Iris Zalaudek}, {and}
  \bibinfo{person}{Harald Kittler}.} \bibinfo{year}{2020}\natexlab{}.
\newblock \showarticletitle{Human-computer collaboration for skin cancer
  recognition}.
\newblock \bibinfo{journal}{\emph{Nature Medicine}} \bibinfo{volume}{26},
  \bibinfo{number}{8} (\bibinfo{year}{2020}), \bibinfo{pages}{1229--1234}.
\newblock


\bibitem[van~der Linden and Hambleton(2013)]%
        {van2013handbook}
\bibfield{author}{\bibinfo{person}{W.J. van~der Linden} {and}
  \bibinfo{person}{R.K. Hambleton}.} \bibinfo{year}{2013}\natexlab{}.
\newblock \bibinfo{booktitle}{\emph{Handbook of Modern Item Response Theory}}.
\newblock \bibinfo{publisher}{Springer, New York}.
\newblock


\bibitem[Vehtari et~al\mbox{.}(2017)]%
        {vehtari2017practical}
\bibfield{author}{\bibinfo{person}{Aki Vehtari}, \bibinfo{person}{Andrew
  Gelman}, {and} \bibinfo{person}{Jonah Gabry}.}
  \bibinfo{year}{2017}\natexlab{}.
\newblock \showarticletitle{Practical Bayesian model evaluation using
  leave-one-out cross-validation and WAIC}.
\newblock \bibinfo{journal}{\emph{Statistics and Computing}}
  \bibinfo{volume}{27} (\bibinfo{year}{2017}), \bibinfo{pages}{1413--1432}.
\newblock


\bibitem[Wang et~al\mbox{.}(2016)]%
        {wang2016deep}
\bibfield{author}{\bibinfo{person}{Dayong Wang}, \bibinfo{person}{Aditya
  Khosla}, \bibinfo{person}{Rishab Gargeya}, \bibinfo{person}{Humayun Irshad},
  {and} \bibinfo{person}{Andrew~H Beck}.} \bibinfo{year}{2016}\natexlab{}.
\newblock \showarticletitle{Deep learning for identifying metastatic breast
  cancer}.
\newblock \bibinfo{journal}{\emph{arXiv preprint arXiv:1606.05718}}
  (\bibinfo{year}{2016}).
\newblock


\bibitem[Wang et~al\mbox{.}(2021)]%
        {wangtowards2021}
\bibfield{author}{\bibinfo{person}{Qiaosi Wang}, \bibinfo{person}{Koustuv
  Saha}, \bibinfo{person}{Eric Gregori}, \bibinfo{person}{David Joyner}, {and}
  \bibinfo{person}{Ashok Goel}.} \bibinfo{year}{2021}\natexlab{}.
\newblock \showarticletitle{Towards Mutual Theory of Mind in Human-AI
  Interaction: How Language Reflects What Students Perceive About a Virtual
  Teaching Assistant}. In \bibinfo{booktitle}{\emph{Proceedings of the 2021 CHI
  Conference on Human Factors in Computing Systems}} (Yokohama, Japan)
  \emph{(\bibinfo{series}{CHI '21})}. \bibinfo{publisher}{Association for
  Computing Machinery}, \bibinfo{address}{New York, NY, USA}, Article
  \bibinfo{articleno}{384}, \bibinfo{numpages}{14}~pages.
\newblock
\showISBNx{9781450380966}
\urldef\tempurl%
\url{https://doi.org/10.1145/3411764.3445645}
\showDOI{\tempurl}


\bibitem[Westby and Riedl(2022)]%
        {Westby2022CollectiveII}
\bibfield{author}{\bibinfo{person}{Samuel Westby} {and}
  \bibinfo{person}{Christoph Riedl}.} \bibinfo{year}{2022}\natexlab{}.
\newblock \showarticletitle{Collective intelligence in human-{AI} teams: A
  {B}ayesian theory of mind approach}.
\newblock \bibinfo{journal}{\emph{ArXiv}}  \bibinfo{volume}{abs/2208.11660}
  (\bibinfo{year}{2022}).
\newblock


\bibitem[Westerman et~al\mbox{.}(2020)]%
        {westermanit2020}
\bibfield{author}{\bibinfo{person}{David Westerman}, \bibinfo{person}{Autumn~P.
  Edwards}, \bibinfo{person}{Chad Edwards}, \bibinfo{person}{Zhenyang Luo},
  {and} \bibinfo{person}{Patric~R. Spence}.} \bibinfo{year}{2020}\natexlab{}.
\newblock \showarticletitle{I-It, I-Thou, I-Robot: The Perceived Humanness of
  AI in Human-Machine Communication}.
\newblock \bibinfo{journal}{\emph{Communication Studies}} \bibinfo{volume}{71},
  \bibinfo{number}{3} (\bibinfo{year}{2020}), \bibinfo{pages}{393--408}.
\newblock
\urldef\tempurl%
\url{https://doi.org/10.1080/10510974.2020.1749683}
\showDOI{\tempurl}
\showeprint{https://doi.org/10.1080/10510974.2020.1749683}


\bibitem[Wilder et~al\mbox{.}(2020)]%
        {wilder2020complement}
\bibfield{author}{\bibinfo{person}{Bryan Wilder}, \bibinfo{person}{Eric
  Horvitz}, {and} \bibinfo{person}{Ece Kamar}.}
  \bibinfo{year}{2020}\natexlab{}.
\newblock \showarticletitle{Learning to complement humans}. In
  \bibinfo{booktitle}{\emph{Proceedings of the Twenty-Ninth International Joint
  Conference on Artificial Intelligence (IJCAI-20)}}.
  \bibinfo{pages}{1526--1533}.
\newblock


\end{thebibliography}

\onecolumn
\appendix
\appendix
\onecolumn

\section{Additional Results and Figures}
\label{sec:additionalresults}

\subsection{From Section \ref{sec:rf1}}
Here we include other metrics (WAIC and LOO scores) for the one-dimensional and multidimensional IRT models. Scores are computed on the log-score scale, i.e., a higher score is better. Note there are no standard errors for the baseline model, as no parameters are learned.

\begin{table}[H]
\centering
\caption{WAIC scores and standard errors of one-dimensional and multidimensional models for other humans and AI agents, across both feedback conditions.}
\begin{tabular}{lll}
           & Humans & AI \\
 \hline\hline
Baseline & -4227.04 & -4144.96 \\
One-dimensional & -3404.6 $\pm$ 28.0    &  -3348.5 $\pm$ 28.7 \\
Multidimensional & -3024.9 $\pm$ 33.7  & -2855.4 $\pm$ 40.3 \\
\end{tabular}
\label{table:multidim0}
\end{table}

\begin{table}[h!]
\centering
\caption{LOO scores and standard errors of one-dimensional and multidimensional models for other humans and AI agents, across both feedback conditions.}
\begin{tabular}{lll}
           & Humans & AI \\
 \hline\hline
 Baseline & -4227.04 & -4144.96 \\
One-dimensional & -3405.1 $\pm$ 28.0    &  -3349.1 $\pm$ 28.7 \\
Multidimensional & -3031.5 $\pm$ 33.9  & -2859.8 $\pm$ 40.4 \\
\end{tabular}
\label{table:multidim1}
\end{table}

\begin{figure*}[h!]
    \centering
    \begin{subfigure}{0.5\textwidth}  
    \centering
        \includegraphics[width=0.9\linewidth]{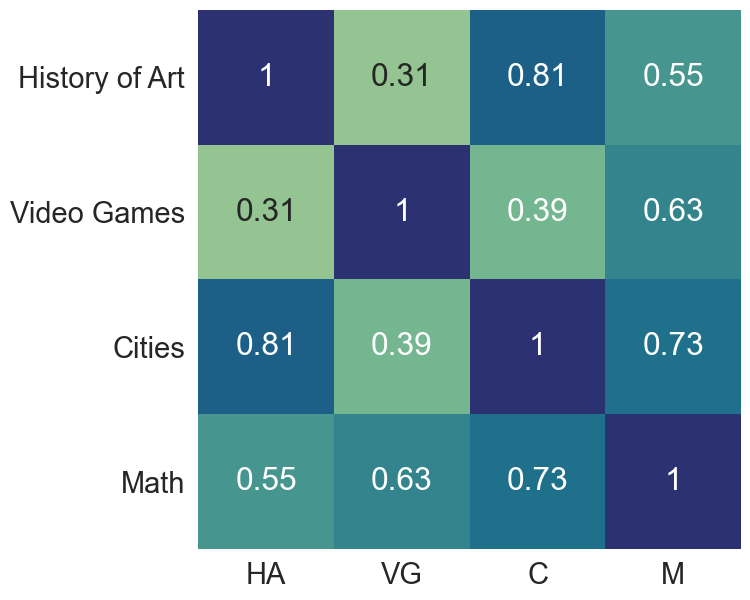}    \caption{AI agents}
        \label{fig:aicorrs2}
    \end{subfigure}%
    \begin{subfigure}{0.5\textwidth}    
    \centering
        \includegraphics[width=0.9\linewidth]{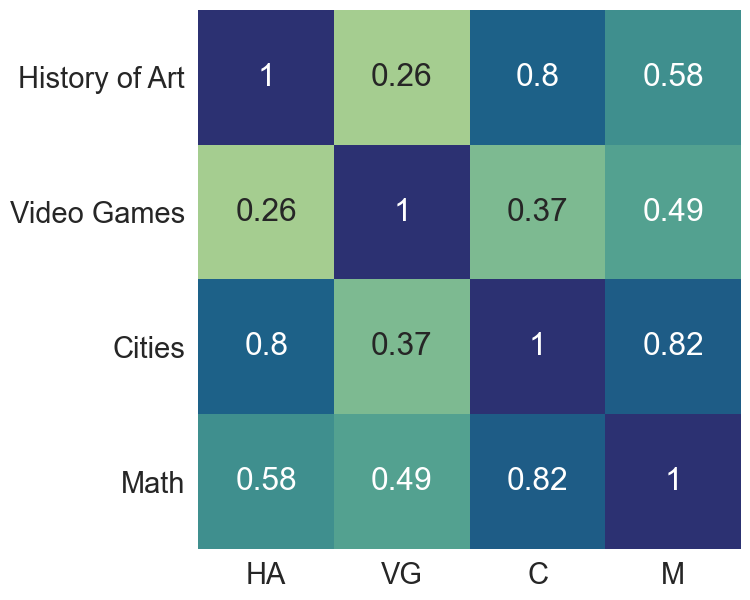}
        \caption{Other humans}
        \label{fig:humancorrs2}
    \end{subfigure}
    \caption{True ability correlations for per-topic performances, split by agent type. These are the latent correlations between abilities, computed from the true other agent performance data.}
\end{figure*}

\begin{figure*}
    \centering
    \begin{subfigure}{0.5\textwidth}
    \centering
        \includegraphics[width=0.6\linewidth]{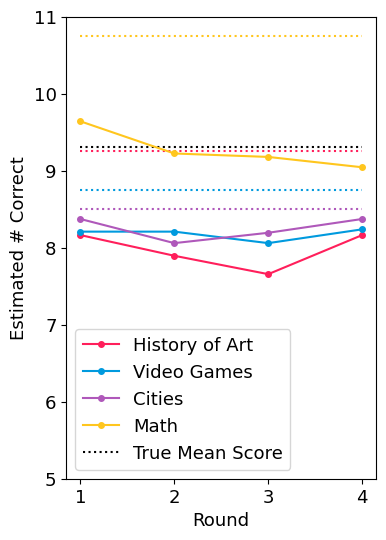}
        \caption{AI agents}
    \end{subfigure}%
    \begin{subfigure}{0.5\textwidth}
    \centering
        \includegraphics[width=0.6\linewidth]{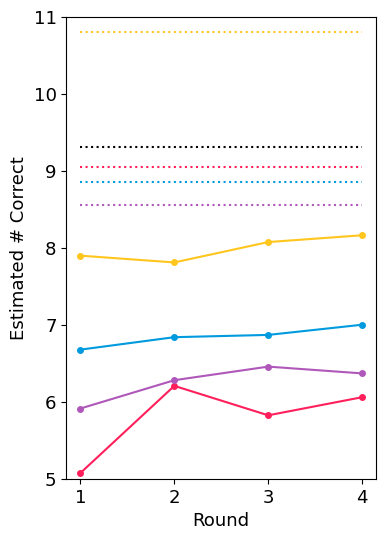}
        \caption{Other humans}
    \end{subfigure}
    \caption{Average other-assessment for high accuracy agents (AI agents in (a), other humans in (b)) in the feedback condition. In both figures, the solid lines plot the average other-assessed performance across all four rounds. The average true performances of the other agent are shown as dotted horizontal lines, with the overall average in black and per-topic averages in their respective colors.}
    \label{fig:time2}
\end{figure*}

\subsection{From Section \ref{sec:rf2}}
Here we include additional scores, averaged over each participant, for each of the three hierarchical models.

\begin{table}[!ht]
\centering
\caption{Held-out next-round log-likelihoods (higher is better) for self differentiation models in the no feedback condition}
\begin{tabular}{lll}
           & Humans & AI \\
 \hline\hline
 Baseline & -2.56 & -2.56 \\
 Undifferentiated &  -2.68 & -5.80 \\
 Differentiated by Ability & \textbf{-1.89}  & \textbf{-1.72} \\
 Fully Differentiated & -2.07 & -1.77  \\
\end{tabular}
\end{table}

\begin{table}[!ht]
\centering
\caption{Average LOO scores for self differentiation  models, across both feedback conditions}
\begin{tabular}{lll}
           & Humans & AI \\
 \hline\hline
 Baseline & -41.04 & -41.04 \\
 Undifferentiated &  -42.58 & -67.89 \\
 Differentiated by Ability & -29.49  & -30.29 \\
 Fully Differentiated & -30.58 & -28.82  \\
\end{tabular}
\end{table}

\begin{table}[!ht]
\centering
\caption{Average WAIC scores for self differentiation  models, across both feedback conditions}
\begin{tabular}{lll}
           & Humans & AI \\
 \hline\hline
 Baseline & -41.04 & -41.04 \\
 Undifferentiated &  -42.58 & -67.89 \\
 Differentiated by Ability & -29.38  & -30.17 \\
 Fully Differentiated & -29.94 & -28.25  \\
\end{tabular}
\end{table}

\section{IRT Model Details and Priors}
\label{sec:irtdetails}
All models were fit using \href{https://mc-stan.org/}{Stan}. The underlying model (\ref{sec:underlying}) was used to model other-assessment data $x_{i,j}^o$ in Section \ref{sec:rf1} as well as true performance $x_{i,j}$ in the top level of the hierarchy in Section \ref{sec:rf2}. The second level of the hierarchy was modeled by the self-assessment model (\ref{sec:selfassessment}). The three other-assessment models are detailed in \ref{sec:otherassessment}.

The underlying models were run with 800 warm-up iterations, 1500 samples, and three chains. Each of the self-assessment and other-assessment models (one for each participant) were run with 600 warm-up iterations, 1000 samples, and 2 chains. These hyperparameters were chosen based on chain convergence plots.

To convert latent scores $\theta_{i,j}$ to discrete scores, we used:
\begin{align*}
&p_{i,j} = \frac{1}{1+\text{exp}(-\theta_{i,j})}\\
&x_{i,j} \sim \text{OrderedProbit}(p_{i,j}, v, \sigma)
\end{align*}
where $v$ is an array of cutoff points for conversion to discrete scores. We used 13 equally-spaced bins between 0 and 1 (converting into a score between 0 and 12, the number of questions in each problem set). 

\subsection{Underlying Model}
\label{sec:underlying}
\begin{equation*}
\begin{aligned}[c]
&\textbf{Multidimensional}\\
&x_{i,j} = f(\lambda_j \cdot \mathbf{a}_i, d_j, \sigma)\\
&\sigma \sim \text{Cauchy}(0,2)\\
&d_j \sim \text{N}(\mu_d, \sigma_d)\\
&\qquad \mu_d \sim \text{N}(0,2)\\
&\qquad \sigma_d \sim \text{Cauchy}(0,5)\\
&\mathbf{a}_i \sim \text{MVN}(\mathbf{0}, \Sigma_L)\\
&\qquad \Sigma_L = L_\text{std} \cdot L_\Omega\\
&\qquad L_\Omega \sim \text{lkj\_corr\_cholesky}(1)\\
&\qquad L_\text{std} \sim \text{N}(0, 2.5)\\
\end{aligned}
\qquad\qquad\qquad\qquad
\begin{aligned}[c]
&\textbf{One-dimensional}\\
&Y_{i,j} = f(a_i, d_j, \sigma)\\
&\sigma \sim \text{Cauchy}(0,2)\\
&d_j \sim \text{N}(\mu_d, \sigma_d)\\
&\qquad \mu_d \sim \text{N}(0,2)\\
&\qquad \sigma_d \sim \text{Cauchy}(0,5)\\
&a_i \sim \text{N}(0,1)\\
&\qquad\\
&\qquad\\
&\qquad\\
\end{aligned}
\end{equation*}

\subsection{Self-Assessment Model}
\label{sec:selfassessment}
\begin{equation*}
\begin{aligned}[c]
&\textbf{Multidimensional}\\
&x^s_{i,j} = f(\lambda_j \cdot \mathbf{a}^s_i, d^s_j, \sigma^s)\\
&\sigma^s \sim \text{Cauchy}(0,2)\\
&d^s_j \sim \text{N}(\gamma \cdot d_j + \Lambda, \sigma_{d,i})\\
&\qquad \gamma \sim \text{N}(0,1)\\
&\qquad \Lambda \sim \text{N}(0,1)\\
&\qquad \sigma_{d,i} \sim \text{Cauchy}(0,2)\\
&a^s_{i,k} \sim \text{N}(a_{i,k}, \sigma_{a,i})\\
&\qquad \sigma_{a,i} \sim \text{Cauchy}(0,2)\\
\end{aligned}
\qquad\qquad\qquad\qquad
\begin{aligned}[c]
&\textbf{One-dimensional}\\
&x^s_{i,j} = f(a^s_i, d^s_j, \sigma^s)\\
&\sigma^s \sim \text{Cauchy}(0,2)\\
&d^s_j \sim \text{N}(\gamma \cdot d_j + \Lambda, \sigma_{d,i})\\
&\qquad \gamma \sim \text{N}(0,1)\\
&\qquad \Lambda \sim \text{N}(0,1)\\
&\qquad \sigma_{d,i} \sim \text{Cauchy}(0,2)\\
&a^s_{i} \sim \text{N}(a_{i}, \sigma_{a,i})\\
&\qquad \sigma_{a,i} \sim \text{Cauchy}(0,2)\\
\end{aligned}
\end{equation*}

\newpage

\subsection{Other-Assessment Models}
\label{sec:otherassessment}
\subsubsection{Undifferentiated}
\begin{equation*}
\begin{aligned}[c]
&\textbf{Multidimensional}\\
&x^o_{i,j} = f(\lambda_j \cdot \mathbf{a}^s_i, d^s_j, \sigma^s)\\
&\text{Input data: } \mathbf{a}^s_i, d^s_j, \sigma^s\\
\end{aligned}
\qquad\qquad\qquad\qquad
\begin{aligned}[c]
&\textbf{One-dimensional}\\
&x^o_{i,j} = f(a^s_i, d^s_j, \sigma^s)\\
&\text{Input data: } a^s_i, d^s_j, \sigma^s\\
\end{aligned}
\end{equation*}

\subsubsection{Differentiated by Ability}
\begin{equation*}
\begin{aligned}[c]
&\textbf{Multidimensional}\\
&x^o_{i,j} = f(\lambda_j \cdot \mathbf{a}^o_i, d^s_j, \sigma^s)\\
&a^o_{i,k} = a^s_{i,k} + \delta_{i,k}\\
&\qquad \delta_{i,k} \sim \text{N}(0,1)\\
&\text{Input data: } \mathbf{a}^s_i, d^s_j, \sigma^s\\
\end{aligned}
\qquad\qquad\qquad\qquad
\begin{aligned}[c]
&\textbf{One-dimensional}\\
&x^o_{i,j} = f(a^o_i, d^s_j, \sigma^s)\\
&a^o_{i} = a^s_{i} + \delta_{i}\\
&\qquad \delta_{i} \sim \text{N}(\mu_{\delta_{i}}, \sigma_\delta)\\
&\qquad \mu_{\delta_{i}} \sim \text{N}(0,1)\\
&\qquad \sigma_\delta \sim \text{Cauchy}(0,2)\\
&\text{Input data: } a^s_i, d^s_j, \sigma^s\\
\end{aligned}
\end{equation*}

\subsubsection{Fully Differentiated}
\begin{equation*}
\begin{aligned}[c]
&\textbf{Multidimensional}\\
&x^o_{i,j} = f(\lambda_j \cdot \mathbf{a}^o_i, d^o_j, \sigma^s)\\
&d^o_j \sim \text{N}(\mu^o_d, \sigma^o_d)\\
&\qquad \mu^o_d \sim \text{N}(0,2)\\
&\qquad \sigma^o_d \sim \text{Cauchy}(0,5)\\
&a^o_{i,k} \sim \text{N}(0,1)\\
&\text{Input data: }\sigma^s\\
\end{aligned}
\qquad\qquad\qquad\qquad
\begin{aligned}[c]
&\textbf{One-dimensional}\\
&x^o_{i,j} = f(a^o_i, d^o_j, \sigma^s)\\
&d^o_j \sim \text{N}(\mu^o_d, \sigma^o_d)\\
&\qquad \mu^o_d \sim \text{N}(0,2)\\
&\qquad \sigma^o_d \sim \text{Cauchy}(0,5)\\
&a^o_i \sim \text{N}(0,1)\\
&\text{Input data: }\sigma^s\\
\end{aligned}
\end{equation*}

\section{Experiment Setup}
\label{sec:experimentsetup}

\begin{figure*}
    \centering
    \includegraphics[width=0.6\linewidth]{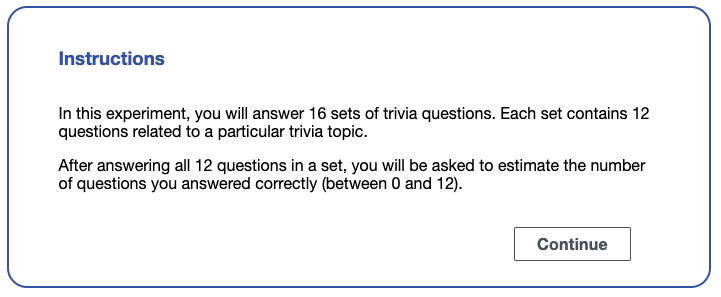} 
    \caption{First page of instructions (for all participants).}
    \label{fig:instructions1}
\end{figure*}

\begin{figure*}
    \centering
    \includegraphics[width=0.6\linewidth]{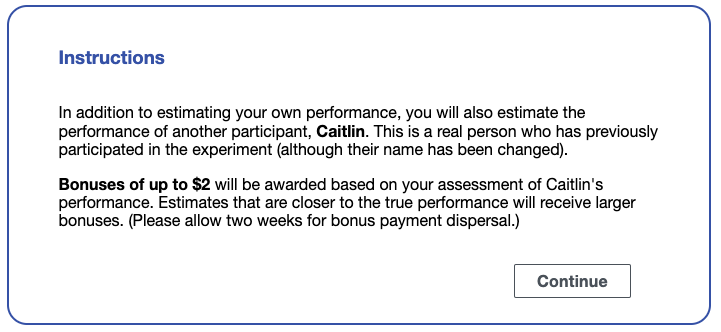} 
    \caption{Second page of instructions shown to participants in the ``other human'' condition.}
    \label{fig:instructions2}
\end{figure*}

\begin{figure*}
    \centering
    \includegraphics[width=0.6\linewidth]{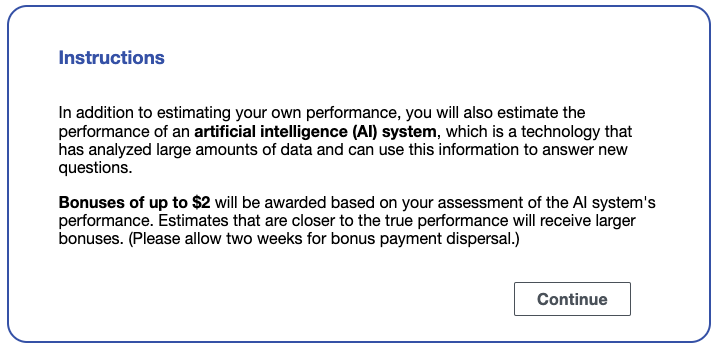} 
    \caption{Second page of instructions shown to participants in the ``AI agent'' condition.}
    \label{fig:instructions3}
\end{figure*}

\begin{figure*}
    \centering
    \includegraphics[width=0.6\linewidth]{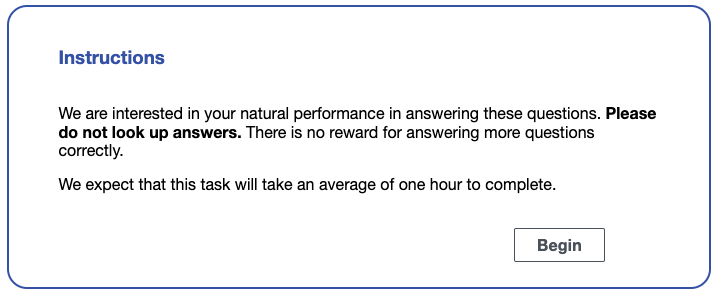} 
    \caption{Third page of instructions (for all participants).}
    \label{fig:instructions4}
\end{figure*}

The instructions provided to participants at the beginning of the experiment are shown in Figures \ref{fig:instructions1} through \ref{fig:instructions4}. The second page of instructions depends on whether the participant was assigned another human or an AI agent to assess.


\begin{figure*}
    \centering
    \includegraphics[width=\linewidth]{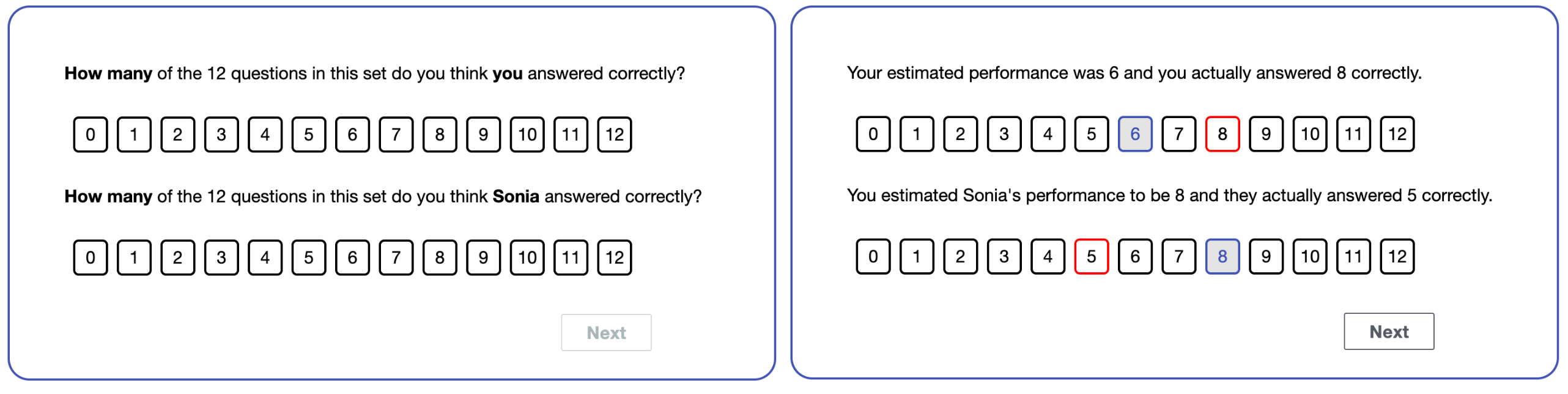} 
    \caption{Example performance estimation questions. In this example, the participant estimates their own performance and the performance of another person, Sonia (left). They receive feedback about their own, and Sonia's, actual performance (right).}
    \label{fig:samplefeedback}
\end{figure*}

To match AI agent performance to that of human performance, we first chose the five highest-accuracy and five lowest-accuracy participants in our pilot study. We then ran six models: three versions of UnifiedQA (base, large, and 3B, see \href{https://github.com/allenai/unifiedqa}{documentation}) and three versions of Zero-shot-CoT (one with GPT3-XL version 1 and method ``zero\_shot,'' two with method ``zero\_shot\_cot'' and GPT3-XL versions 1 and 3, respectively; see \href{https://github.com/kojima-takeshi188/zero_shot_cot}{documentation}). Testing this wide variety of models enabled us to match accuracy relatively closely to the humans. The models used for each topic are shown in Table \ref{table:aimodels}.

\begin{table}[h!]
\centering
\caption{Models used for AI Agent}
\begin{tabular}{llll|cccccc}
           & Topic & Human & Model & UQA (base) & UQA (large) & UQA (3B) & ZS (001) & ZSC (001) & ZSC (003) \\
 \hline\hline
High accuracy & Art & 75.4 & 77.1 & & & & \checkmark & & \\
& Video Games & 73.8  & 72.9 & & & & & \checkmark &   \\
& Cities & 71.3  & 70.8 & & & \checkmark & & &   \\
& Math & 90.0  & 89.6 & & & & & &  \checkmark \\
\hline
Low accuracy & Art & 30.0 & 29.2 & \checkmark & & & & & \\
& Video Games & 51.3  & 52.1 & & \checkmark & & & &   \\
& Cities & 33.8 & 31.3 & \checkmark & & & & &   \\
& Math & 60.0 & 62.5 & & & & \checkmark & &   \\
\end{tabular}
\label{table:aimodels}
\end{table}

\end{document}